\documentclass{article}

\usepackage[english]{babel}
\usepackage{authblk}

\usepackage[letterpaper,top=2cm,bottom=2cm,left=3cm,right=3cm,marginparwidth=1.75cm]{geometry}

\usepackage{amsmath}
\usepackage{graphicx}
\usepackage[round]{natbib}
\usepackage{xcolor}
\definecolor{darkblue}{rgb}{0.0, 0.0, 0.55} 

\usepackage[colorlinks=true,
            linkcolor=darkblue,
            urlcolor=blue,
            citecolor=darkblue,
            anchorcolor=darkblue]{hyperref}
\usepackage[T1]{fontenc}
\usepackage{amsmath}
\usepackage{amssymb}
\usepackage{mathtools}
\usepackage{amsthm}
\usepackage{microtype}
\usepackage{graphicx}
\usepackage{subcaption}
\usepackage{booktabs} 
\usepackage{bm}
\usepackage{multirow}

\title{SIGMA-PPG: Statistical-prior Informed Generative 
    \texorpdfstring{\\}{ } Masking Architecture for PPG Foundation Model}
    
\author[1]{Zongheng Guo}
\author[1,2]{Tao Chen\thanks{Corresponding author: \texttt{chentao98@zju.edu.cn}}}
\author[3]{Yang Jiao}
\author[3]{Yi Pan}
\author[4]{Xiao Hu}
\author[1]{Manuela Ferrario}

\affil[1]{Department of Electronics, Information and Bioengineering, Politecnico di Milano, Milan, Italy}
\affil[2]{State Key Laboratory of Industrial Control Technology, Zhejiang University, Hangzhou, China}
\affil[3]{Shenzhen Institute of Advanced Technology, Chinese Academy of Sciences, Shenzhen, China}
\affil[4]{Nell Hodgson Woodruff School of Nursing, Emory University, Atlanta, USA}

\date{}

\begin{document}
\maketitle

\begin{abstract}
Current foundation model for photoplethysmography (PPG) signals is challenged by the intrinsic redundancy and noise of the signal. Standard masked modeling often yields trivial solutions while contrastive methods lack morphological precision. To address these limitations, we propose a Statistical-prior Informed Generative Masking Architecture (SIGMA-PPG), a generative foundation model featuring a Prior-Guided Adversarial Masking mechanism, where a reinforcement learning-driven teacher leverages statistical priors to create challenging learning paths that prevent overfitting to noise. We also incorporate a semantic consistency constraint via vector quantization to ensure that physiologically identical waveforms—even those altered by recording artifacts or minor perturbations—map to shared indices. This enhances codebook semantic density and eliminates redundant feature structures. Pre-trained on over 120,000 hours of data, SIGMA-PPG achieves superior average performance compared to five state-of-the-art baselines across 12 diverse downstream tasks. The code is available at \url{https://github.com/ZonghengGuo/SigmaPPG}.
\end{abstract}

\begin{figure*}[t]
    \centering
    \includegraphics[width=\textwidth]{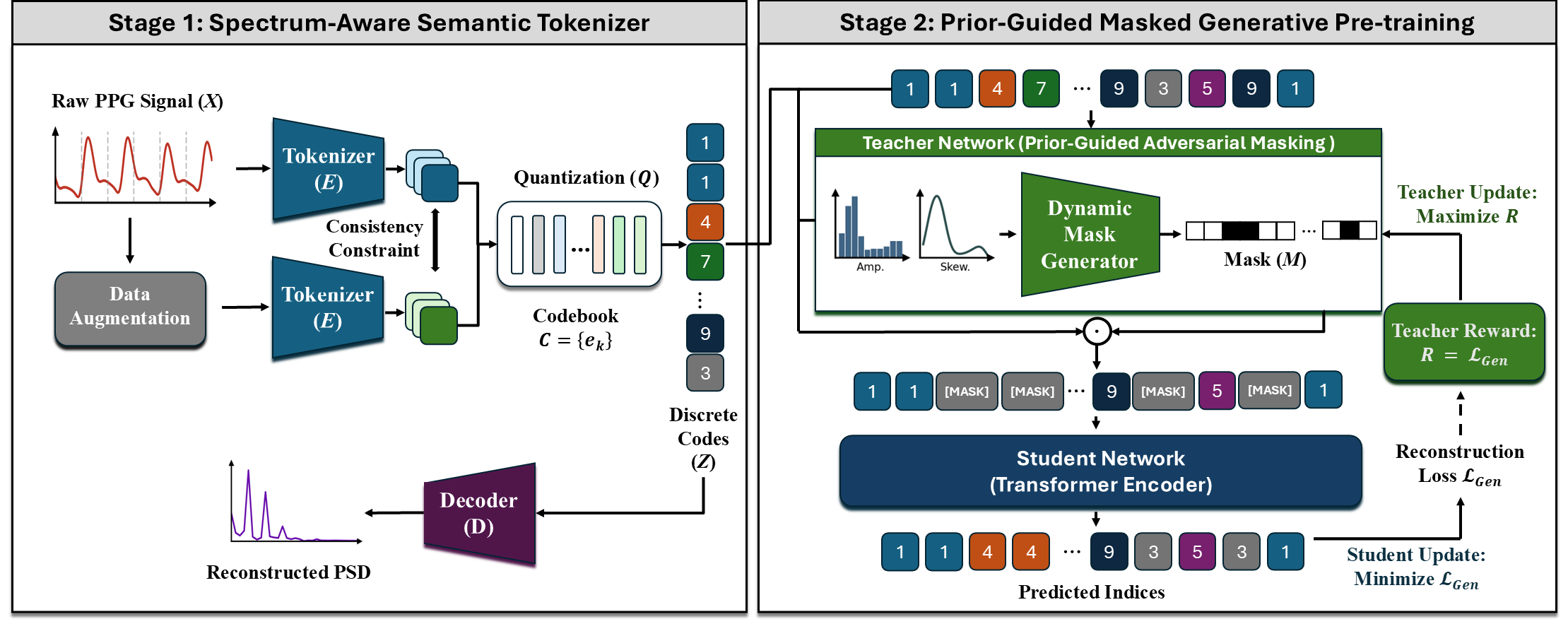}
    \caption{\textbf{Overview of the SIGMA-PPG model framework}. The architecture consists of two cascading stages: (1) Stage 1: Spectrum-Aware Semantic Tokenizer. At this stage, a VQ-VAE is used to map continuous raw PPG signals into discrete semantic tokens. A Power Spectral Density (PSD) reconstruction objective is employed to capture physiological frequency characteristics. Furthermore, a semantic consistency constraint is introduced by a vector quantization to ensure that physiologically identical waveforms, even those perturbed by artifacts or noise, map to consistent codebook indices. (2) Stage 2: Prior-Guided Masked Generative Pre-training. This stage employs a Reinforcement Learning-driven Teacher-Student framework to replace standard random masking. The Teacher network  employs statistical priors (amplitude and skewness) to construct a dynamic curriculum and generate challenging masking policies, i.e., Prior-Guided Adversarial Masking. This mechanism effectively guides the Student network (a Transformer Encoder) to avoid noise overfitting and capture global morphological dependencies.}
    \label{fig:overview}
\end{figure*}

\section{Introduction}
Photoplethysmography (PPG) has become a cornerstone modality for continuous, non-invasive health monitoring. Widely adopted in both clinical environments, such as Intensive Care Units (ICUs), and consumer wearable devices including smartwatches and smart rings, PPG provides to be highly informative for non-invasive cardiovascular monitoring. PPG signal captures rich physiological dynamics, underpinning a wide spectrum of applications ranging from heart rate \citep{reiss2019deep} and oxygen saturation (SpO2) monitoring \citep{bagha2011real} to cuff-less blood pressure estimation \citep{kurylyak2013neural} and emotion recognition \citep{udovivcic2017wearable}. Despite the broad adoption of deep learning (DL), DL in this domain remains constrained by signal-intrinsic limitations, including low signal-to-noise ratio (SNR) and strong vulnerability to motion-induced artifacts. Models trained on limited labeled datasets often overfit to specific noise profiles or simple periodic patterns, resulting in poor generalization when deployed in different contexts or for different purposes. The PPG signal characteristics can vary substantially due to factors unrelated to physiology, such as motion artifacts, sensor displacement, contact pressure, ambient light, or device-specific characteristics. This vulnerability may obscure or dominate the underlying physiological dynamics\citep{ghorbani2023self}. To overcome the limitations imposed by scarce labeled data and to learn representations that faithfully capture the latent manifold of physiological signals under such variability, the research community has increasingly adopted self-supervised learning (SSL) on large-scale unlabeled datasets.\citep{ding2024self}.

Pioneering models such as PaPaGei \citep{pillai2024papagei}, Pulse-PPG \citep{saha2025pulse}, and AnyPPG \citep{nie2025anyppg} have successfully utilized large-scale datasets, predominantly adopting the contrastive learning paradigm to learn invariant features via data augmentation. Although effective for learning robust features, these discriminative frameworks exhibit intrinsic limitations when applied to physiological signals. Contrastive learning promotes global invariance for sample discrimination, which can unintentionally suppress subtle but clinically meaningful morphological details, such as variations in the dicrotic notch or in the slope, treating them as intra-sample noise. 
Consequently, these models often yield coarse-grained representations that lack the generative understanding of the signal's fine-grained temporal evolution, limiting their transferability to downstream tasks demanding precise waveform reconstruction or dense parameter estimation \citep{atienza2024contrastive} \citep{li2023ti}.

Drawing inspiration from the success of Large Language Model (LLM) paradigms in physiological domains, exemplified by models like LaBraM \citep{jiang2024labram} for EEG signals and AnyECG \citep{wang2024anyecg} for ECG recordings, we explore a generative masked modeling approach for PPG signals. While promising, directly transposing this discrete tokenization and masking paradigm to PPG signal entails unique challenges, primarily due to the signal's intrinsic redundancy and discretization instability. First, the highly quasi-periodicity within typical PPG signals renders standard random masking trivial. Unlike in natural language where context is complex, in this case models can easily minimize training loss by simply copying patterns from adjacent cardiac cycles or performing local interpolation without learning genuine hemodynamic semantics. Second, the noise-sensitive nature of PPG conflicts with the discrete boundaries of vector quantization (VQ). In the presence of ubiquitous baseline wander, physiologically identical waveforms often suffer from a topology mismatch, being mapped to different codebook indices \citep{chen2025does}. This instability in the discretization process fractures the semantic continuity of the latent space, causing the model to encode artifacts as distinct semantic tokens. To overcome these limitations and effectively unleash the potential of the LLM paradigm, we propose a Statistical-prior Informed Generative Masking architecture for PPG signal (SIGMA-PPG).

First, we propose a Prior-Guided Adversarial Masking mechanism. To address the ineffectiveness  of random masking on PPG signals, we formulate mask generation as a Teacher-Student game driven by Reinforcement Learning (RL). In this framework, Teacher's objective is to generate the most challenging masking patterns possible in order to maximize the Student's reconstruction loss. This adversarial game establishes a dynamic pattern learning process that forces the Student to abandon simple local interpolation in favor of understanding the signal's morphological structure globally. However, unconstrained adversarial training can lead to a "degenerate solution" trap: the Teacher can realize that completely random spike artifacts, such those resulting from signal loss, result in the highest reconstruction error, as these non-stationary noise segments are mathematically unpredictable. However, this contributes nothing to learning meaningful physiological features. To address this issue, we have explicitly incorporated the skewness and amplitude of the PPG signal as statistical biases directly into the Teacher’s policy logits. These statistics effectively distinguish between waveforms that are rich in physiological information and those that are meaningless noise. By modulating the sampling probability distribution with this prior guidance, we adjust the Teacher's optimization trajectory, directing the generated masks to focus on critical morphological regions (e.g., systolic peaks), thereby guiding the Student to learn truly robust physiological representations.

Secondly, we introduce a semantic consistency constraint. In order to mitigate the low SNR and prevent the model from encoding nuisance variations (e.g. due to motion artifacts, slight noise or baseline wander) as distinct features, we utilize Vector Quantization (VQ) combined with a consistency loss. This ensures that perturbed versions of the same biological signal map to the same codebook indices in the discrete space. This mechanism addresses the common issues of redundancy and discontinuity common in PPG signals by compelling the model to disregard subtle morphological discrepancies and group physically similar signals into the same semantic cluster.

Thirdly, we pre-trained our model using a large scale public dataset comprising more than 120,000 hours of PPG signal recordings and fine-tuned it using 12 diverse tasks spanning 6 downstream datasets, covering both regression and classification. Comprehensive comparisons with five state-of-the-art PPG foundation models demonstrate that SIGMA-PPG has strong potential to serve as a unified and robust backbone for next-generation medical AI and wearable health applications.

\section{Method}

As illustrated in Figure 1, the SIGMA-PPG architecture consists of two successive steps.

Stage 1: Spectrum-Aware Semantic Tokenizer.
Given a normalized single-channel PPG signal $x$ of $L$ samples, we first partition it into a sequence of $N$ non-overlapping patches,  $\mathcal{X} = \{x_1, x_2, \dots, x_N\}$, where each patch $x_i$ of $T$ samples acts as a local semantic unit.
A tokenizer network ($E$) maps each patch to a discrete token using a learnable codebook $\mathcal{C}=\{e_k\}$ so that the power spectral density (PSD) of the original signal can be reconstructed, preserving frequency information and semantic consistency is enforced \citep{jiang2024labram} \citep{guo2025qualityfm}. The input patches are then mapped to a sequence of discrete semantic tokens $\mathcal{Z} = \{z_1, z_2, \dots, z_N\}$. Global statistical properties of the signal, specifically amplitude and skewness, are computed. These serve as prior knowledge ($S_{\text{prior}}$) that helps guide later stages of the model.

Stage 2: Prior-Guided Generative Pre-training. The Teacher model dynamically constructs a mask $M$ based on a Reinforcement Learning (RL) framework and $S_{\text{prior}}$. The Student model then attempts to recover the indices of the masked tokens within the codebook $\mathcal{C}$, yielding the reconstructed sequence $\tilde{\mathcal{Z}}$ \citep{vaswani2017attention}.

\subsection{Stage 1: Spectrum-Aware Neural Tokenizer}
To transform continuous PPG signals into a discrete semantic sequence, we train an enhanced Vector Quantized Variational Autoencoder (VQ-VAE) \citep{van2017neural}.

\subsubsection{Vector Quantization}
We partition a normalized single-channel PPG signal $x \in \mathbb{R}^{L}$,specifically, we adopt a number of samples $L$ corresponding to a 4-minute temporal window as validated in Appendix \ref{sec:window}), into a sequence of $N$ non-overlapping patches $\mathcal{X}= \{x_1, x_2, ..., x_N\}$, where each patch $x_i \in \mathbb{R}^{T}$ represents a one-second temporal window. An encoder $E(\cdot)$ projects each patch $x_i$ into a latent vector $h_i = E(x_i) \in \mathbb{R}^{D}$. 
Subsequently, we discretize these vectors using a learnable codebook $\mathcal{C} = \{e_k\}\in \mathbb{R}^{K \times D}$. The quantizer $Q(\cdot)$ maps each latent vector $h_i$ to its nearest neighbor in the codebook $\mathcal{C}$ as follows:
\begin{equation}
z_i = \mathop{\arg\min}_{k \in \{1,\cdots, K\}} \| h_i - e_k \|_2
\end{equation}
where $z_i$ denotes the discrete token index, therefore we can obtain the discrete semantic tokens $\mathcal{Z} = \{z_1, z_2, \dots, z_N\}$. To enable backpropagation through the non-differentiable quantization operation, we employ the Straight-Through Estimator (STE), copying gradients from the decoder directly to the encoder \citep{bengio2013estimating}.

\subsubsection{Spectral reconstruction}

Standard time-domain reconstruction objectives (e.g., MSE) often cause VQ-VAEs to prioritize high-frequency noise over the intrinsic quasi-periodic characteristics of PPG signals. To address this, our decoder $D(\cdot)$ is designed to reconstruct the amplitude spectrum. 
Specifically, we minimize the $L_1$ distance between the decoder's output and the magnitude of the input's Fourier amplitude spectrum, i.e. the magnitude of the Discrete Time Fourier (DTF) coefficients $|X_i[k]|$:
\begin{equation}
\mathcal{L}_{\text{Spec}} = \| |X_i[k]| - D(e_{z_i}) \|_1
\end{equation}
By targeting the amplitude spectrum, the codebook is forced to capture dominant physiological frequencies (e.g., heart rate components) while ignoring phase-sensitive random noise. The ablation study regarding spectrum reconstruction is provided in Appendix \ref{sec:spectrum}.

\subsubsection{Semantic consistency constraint}

To further strengthen the codebook robustness against minor morphological deformations, we introduce a semantic consistency loss. For a given patch $x_i$, we apply stochastic augmentation $\mathcal{A}(\cdot)$, including scaling and Gaussian noise addition, to generate an augmented view $x'_i = \mathcal{A}(x_i)$. We enforce consistency between the latent representations of the original and augmented views prior to quantization:
\begin{equation}
\mathcal{L}_{\text{Con}} = \| \text{sg}[E(x_i)] - E(x'_i) \|_2^2
\end{equation}
where $\text{sg}[\cdot]$ denotes the stop-gradient operator \citep{chen2021exploring}. Consequently, the total training objective for the first stage is formulated as:
\begin{equation}
\mathcal{L}_{\text{Tokenizer}} = \mathcal{L}_{\text{Spec}} + \mathcal{L}_{\text{VQ}} + \mathcal{L}_{\text{Con}}
\end{equation}
where $\mathcal{L}_{\text{VQ}}$ represents the standard codebook commitment loss. The ablation experiment and geometric validation of semantic consistency constraint are provided in Appendix~\ref{sec:consistency loss} and \ref{app:consistency_validation}, respectively.

\subsection{Stage 2: Prior-Guided Generative Pre-training}

In the second stage, we freeze the tokenizer and utilize the generated discrete token sequence $Z = \{z_1, ..., z_N\}$ to train a transformer-based generative learner. To construct an effective masking curriculum, we propose a Teacher-Student framework that integrates statistical priors.

\subsubsection{Formulation of statistical priors}

To prevent the model from learning invalid features in regions dominated by pure noise or flat signal, we design a composite scoring mechanism based on amplitude stability and morphological skewness. These act as prior knowledge $S_i$ to evaluate the physiological information density of each patch $x_i$. Examples of how statistical priors are used in this work are provided in Appendix~\ref{sec:knowledge}.

\paragraph{Amplitude stability score ($S_{\text{amp}}$).}
The amplitude of physiological signals should fluctuate within a reasonable dynamic range. For example, extremely low amplitudes typically indicate poor sensor contact or signal loss, while exaggerated amplitudes suggest severe motion artifacts. We design a scoring function based on a trapezoidal flat-top window, combining relative stability and absolute validity.
First, we compute the set of standard deviations for all patches within the current signal $x$, $\boldsymbol{\sigma} = \{\sigma_1, \dots, \sigma_N\}$.

\begin{itemize}
    \item \textbf{Relative Stability:} to detect abrupt changes, we utilize the Median Absolute Deviation (MAD) \citep{leys2013detecting} to compute a modified Z-score, penalizing outliers that deviate from the global distribution:
    \begin{equation}
    q_i = 0.6745 * \frac{\sigma_i - \text{median}(\boldsymbol{\sigma})}{\text{MAD}(\boldsymbol{\sigma})}, \quad S_{\text{rel}, i} = e^{-0.2 q_i^2}
    \end{equation}

    \item \textbf{Absolute Validity:} we define a broad ``valid range'' $[\sigma_{\min}, \sigma_{\max}]$ using Sigmoid gating functions to enforce lower and upper bounds:
    \begin{equation}
    S_{\text{abs}, i} = \underbrace{\frac{1}{1 + e^{-k_{\text{rise}}(\sigma_i - \sigma_{\min})}}}_{\text{Lower-bound Gate}} \cdot \underbrace{\frac{1}{1 + e^{k_{\text{fall}}(\sigma_i - \sigma_{\max})}}}_{\text{Upper-bound Gate}}
    \end{equation}
\end{itemize}
The range $[0.05, 2.0]$, $k_{\text{rise}}=50$  and $k_{\text{fall}}=5$ are used.. The final amplitude score is the product of the two: $S_{\text{amp}, i} = S_{\text{rel}, i} \cdot S_{\text{abs}, i}$. This solution ensures that high-quality signals of large amplitude are not misclassified as artifacts while effectively filtering extremely noisy segments, flat lines or segments without signal (e.g. flat lines).

\paragraph{Morphological skewness score ($S_{\text{skew}}$).}
Genuine pulse waves typically exhibit significant non-zero skewness due to the signal differences in the systolic and diastolic phase \citep{elgendi2012analysis}, in contrast Gaussian white noise or baseline drift tends to follow a symmetric distribution. We leverage skewness as a feature to distinguish valid waveforms from noise:
\begin{equation}
S_{\text{skew}, i} = \tanh \left( \left| \frac{\frac{1}{T} \sum_{t=1}^{T} (x_{i,t} - \bar{x}_i)^3}{\left( \frac{1}{T} \sum_{t=1}^{T} (x_{i,t} - \bar{x}_i)^2 \right)^{3/2}} \right| \right)
\end{equation}
where $x_{i,t}$ represents the signal value at time point $t$ within patch $x_i$, and $\bar{x}_i$ is the mean of the patch. The hyperbolic tangent function maps the absolute skewness onto the interval $[0, 1)$; higher values indicate higher skewness and therefore a signal with a physiological meaning.

The final statistical prior score $S_{\text{prior}}$ is calculated as a weighted sum:
\begin{equation}
S_{\text{prior}, i} = (1 - \beta) \cdot S_{\text{amp}, i} + \beta \cdot S_{\text{skew}, i}
\end{equation}
where $\beta \in [0, 1]$ is a hyperparameter that balances the contribution of amplitude consistency and skewness. Specifically, we set $\beta = 0.5$ in this work. The sensitivity analysis of $\beta$ is provided in \ref{sec:beta}.

\subsubsection{Prior-Guided Adversarial Masking}

We model mask generation as a reinforcement learning process, employing a collaborative Teacher-Student framework.

\textbf{Reinforcement Learning Formulation.} 
We define the mask generation process as a one-step Markov Decision Process (MDP). The \textbf{State $S$} is defined as the sequence of raw PPG patches $\mathcal{X} = \{x_1, \dots, x_N\}$, providing the Teacher with the full morphological context. Based on this state, the \textbf{Action $A$} corresponds to a binary mask vector $M \in \{0, 1\}^N$, where $M_i=1$ indicates the $i$-th token is masked. To prevent information leakage or complete occlusion, we enforce a fixed masking ratio $r \in [0,1]$ (we set to $r=0.5$ in the experiments). Accordingly, the number of masked patches is defined as $k=\lfloor r\cdot N\rfloor$. The reward $R$ is calculated as the Student's reconstruction loss on the masked tokens ($R = \mathcal{L}_{\text{Gen}}$), which the Teacher aims to maximize.

\textbf{The Student Network.} 
Our backbone is a bidirectional transformer encoder. The model aims to recover the original tokens $z_i$ given the masked sequence $\tilde{Z}$ obtained by the mask $M$ generated by the Teacher. The objective function is the negative log-likelihood of the masked tokens:
\begin{equation}
\mathcal{L}_{\text{Gen}} = -\sum_{i=1}^{N} M_i \log p(z_i | \tilde{Z})
\end{equation}
\textbf{The Teacher Network and Prior-Guided Sampling.} 
The Teacher network $\mathcal{T}$ is designed to generate the most challenging masks $M$. To prevent the Teacher from generating random masks, which lead to uninformative solutions, we explicitly include the aforementioned statistical priors as a Prior Bias into the Teacher's decision-making process \citep{shi2022adversarial}.

The Teacher outputs not normalized logits $L_{\text{Teacher}} \in \mathbb{R}^N$. Before sampling the mask, we add the standardized prior scores into the logits:
\begin{equation}
\text{Bias}_i = \frac{S_{\text{prior}, i} - \mu_S}{\sigma_S}, \quad L_{\text{Final}, i} = L_{\text{Teacher}, i} + \alpha \cdot \text{Bias}_i
\end{equation}
where $\alpha$ is a scaling coefficient, which is set to 2.0 in our experiments. We then compute the probability distribution over all patches:
\begin{equation}
P = \text{Softmax}(L_{\text{Final}}) = \frac{\exp(L_{\text{Final}, i})}{\sum_{j=1}^{N} \exp(L_{\text{Final}, j})}
\end{equation}

\paragraph{Stochastic top-k sampling via Gumbel Perturbation.}
To strictly enforce the constraint of masking exactly $k$ patches without replacement, while maintaining a tractable sampling mechanism, we employ the Gumbel-Top-k trick~\citep{kool2019stochastic}. Unlike standard independent Bernoulli sampling or simple multinomial approximations, this method provides a mathematically rigorous formulation for sampling from a categorical distribution without replacement, consistent with the Plackett-Luce model.

Formally, given the not normalized logits $L_{\text{Final}} \in \mathbb{R}^N$ from the Teacher, we first perturb them with independent and identically distributed (i.i.d.) Gumbel noise:
\begin{equation}
    G_i \sim \text{Gumbel}(0, 1), \quad \forall i \in \{1, \dots, N\}
\end{equation}
\begin{equation}
    \tilde{y}_i = L_{\text{Final}, i} + G_i
\end{equation}
The binary mask $M$ is then constructed by selecting the indices corresponding to the $k$ largest perturbed values in $\tilde{y}$:
\begin{equation}
    M_i = \begin{cases} 
        1 & \text{if } \tilde{y}_i \in \text{top-}k(\{\tilde{y}_1, \dots, \tilde{y}_N\}) \\
        0 & \text{otherwise}
    \end{cases}
\end{equation}
This perturbation mechanism implicitly defines a policy $\pi_\theta(M | \mathcal{X})$ equivalent to sequential sampling without replacement. By leveraging this formulation, we ensure that the generated masks strictly adhere to the constraint of exactly $k$ masked patches while accurately reflecting the distributional preferences learned by the Teacher.

\textbf{Handling non-differentiable constraints.}
The multinomial sampling process itself is stochastic and non-differentiable. However, the Teacher is optimized via policy gradient methods, which only require $\nabla_\theta \log \pi_\theta(M)$, a quantity that is differentiable with respect to $\theta$ despite the discrete sampling. Specifically, $\log P_{id_j}$ depends on $L_{\text{Final}}$ through the Softmax, enabling backpropagation.
After sampling, we apply span constraint to limit the number of consecutive masked patches (e.g., $\le 5$) via a deterministic rule to prevent creating overly long gaps that would disrupt local temporal dependencies.
Specifically, these constraints are applied \textit{after} the multinomial sampling and \textit{do not participate in gradient computation}. The Teacher is trained solely based on the reward signal from the final masked series of patches, making the policy gradient estimator unbiased with respect to the stochastic sampling process. Since these constraints function as part of the environment dynamics, the Teacher optimizes its policy within this constrained action space by observing the reward feedback derived from the final valid masks.

\textbf{Teacher update via REINFORCE.}
This mechanism creates a so-called Physiology-Aware Curriculum \citep{bengio2009curriculum}: in the early training stages, the Prior Bias guides the Teacher to preferentially mask regions with moderate amplitude and significant skewness, i.e., high-quality systolic peaks, forcing the Student to focus on core physiological features. As training progresses, the Teacher gradually explores more complex masking patterns to enhance representation robustness. The visualization of this training dynamics and the curriculum formation is provided in Appendix \ref{sec:training loss}.

The Teacher's update objective is to maximize the expected Student's loss $\mathbb{E}_{M \sim \pi_\theta}[\mathcal{L}_{\text{Gen}}]$ using the REINFORCE algorithm \citep{williams1992simple}. To reduce variance, we subtract a baseline $b$, which is the batch-averaged loss, from the reward:
\begin{equation}
\nabla_\theta J_{\text{Teacher}} = \mathbb{E}_{M \sim \pi_\theta} \left[ (\mathcal{L}_{\text{Gen}}(M) - b) \cdot \nabla_\theta \log \pi_\theta(M | \mathcal{X}) \right]
\end{equation}
where $b = \frac{1}{B} \sum_{j=1}^B \mathcal{L}_{\text{Gen}}^{(j)}$ represents the average difficulty of the current batch. In practice, we approximate this expectation using a single Monte Carlo sample per input instance within the mini-batch. The gradient is then averaged over the entire batch:
\begin{equation}
\nabla_\theta J_{\text{Teacher}} \approx \frac{1}{B} \sum_{j=1}^{B} [ (\mathcal{L}_{\text{Gen}}(M^{(j)}) - b) \cdot \nabla_\theta \log \pi_\theta(M^{(j)} | \mathcal{X}^{(j)})]
\end{equation}
Given the large batch size used in pre-training ($B=4096$), this batch-averaged estimation effectively minimizes the variance of the gradient estimator, stabilizing the adversarial dynamics and preventing reward drift.

\section{Experiments}

We pre-trained the SIGMA-PPG model using PPG signals from two clinical datasets (VitalDB and MIMIC-III) that are different from those used for downstream benchmarks. Detailed descriptions of pre-training data and settings are provided in Appendix~\ref{sec:pretraining settings}.

\subsection{Downstream Tasks}
Table~\ref{tab:dataset} presents the datasets and the tasks used in this paper. More detailed information regarding the datasets is provided in Appendix~\ref{sec:downstream}.

\begin{table}[h]
    \centering
    \caption{Summary of the datasets and corresponding physiological tasks employed in the downstream evaluation. R denotes Regression tasks; B denotes Binary Classification tasks; M denotes Multi-class Classification tasks.}
    \label{tab:dataset}
    
    \resizebox{\columnwidth}{!}{%
        \begin{tabular}{@{}llcc@{}}
        \toprule
        \textbf{Dataset} & \textbf{Computed variable (Task Type)} & \textbf{Sampl. Frequency (Hz)} & \textbf{\#Subj.} \\ \midrule
        BIDMC & Respiratory Rate (R) &125 & 53 \\
        \citep{pimentel2016toward}      & Heart Rate (R) &125& 53 \\
              & SpO2 (R) &125& 53 \\
        PPG-BP & Systolic Blood Pressure (R) &1000 & 219 \\
        \citep{liang2018new}       & Diastolic Blood Pressure (R) &1000 & 219 \\
               & Average Heart Rate (R)&1000 & 219 \\
               & Hypertension (B)&1000 & 219 \\
        WESAD & Stress (B)& 64 & 15 \\
        \citep{schmidt2018introducing}      & Emotion (B)& 64 & 15 \\ 
        Stanford & Signal Quality (M) & 128 & 148 \\
        \citep{torres2020multi} \\
        DaLiA  & Activity (M) & 64 & 15 \\
        \citep{reiss2019ppg} \\
        Real-World PPG & Human Identification (M) & 50 & 35 \\
        \citep{siam2019real} \\
              
              \bottomrule
        \end{tabular}%
    }
\end{table}

\begin{table*}[t]
\centering
\caption{\textbf{Full Fine-Tuning Performance.} Comparison of the proposed SIGMA-PPG models against state-of-the-art baselines under the full fine-tuning setting. The best results are highlighted in \textbf{bold}, the second best are \underline{underlined}.}
\label{tab:main_results_clean}

\resizebox{\linewidth}{!}{
    \setlength{\tabcolsep}{4pt} 
    \begin{tabular}{l|lllll|ll}
    \toprule
    \multirow{2}{*}{\textbf{Classification - AUC ($\uparrow$)}} 
    & \textbf{PAPAGEI-S} \scriptsize{(5M)} & \textbf{PAPAGEI-P} \scriptsize{(5M)} & \textbf{Pulse-PPG} \scriptsize{(28M)} 
    & \textbf{AnyPPG}\scriptsize{(5M)} & \textbf{GPT-PPG}\scriptsize{(19M)}
    & \textbf{SIGMA-PPG} \scriptsize{(5M)} & \textbf{SIGMA-PPG} \scriptsize{(30M)} \\
    
     & \citep{pillai2024papagei} & \citep{pillai2024papagei} & \citep{saha2025pulse}
     & \citep{nie2025anyppg} & \citep{chen2025gpt}
     & \multicolumn{1}{c}{(Ours)} & \multicolumn{1}{c}{(Ours)} \\
    \midrule

    Stress      & 0.9601$\pm$0.01 & 0.9568$\pm$0.00  & 0.9852$\pm$0.02  & 0.9815$\pm$0.00 & \underline{0.9899$\pm$0.01} & 0.9256$\pm$0.03 & \textbf{0.9994$\pm$}\textbf{0.00} \\
    Affect      & 0.7194$\pm$0.01  & 0.6945$\pm$0.00 & 0.7630$\pm$0.03  & 0.8407$\pm$0.01  & \underline{0.8516$\pm$0.03} & 0.7145$\pm$0.05  & \textbf{0.8620}$\pm$\textbf{0.04} \\
    Hypertension & 0.5149$\pm$0.10 & 0.5593$\pm$0.07  & \textbf{0.8258}$\pm$\textbf{0.11}  & 0.5085$\pm$0.26 & 0.7496$\pm$0.13  & 0.7045$\pm$0.10 & \underline{0.7619$\pm$0.11}  \\
    Signal Quality & 0.8923$\pm$0.05 & 0.9162$\pm$0.03 & 0.9734$\pm$0.01  & 0.9523$\pm$0.01 & 0.9587$\pm$0.02 & \underline{0.9821$\pm$0.00} & \textbf{0.9907}$\pm$\textbf{0.00}  \\
    Activity & 0.8533$\pm$0.00 & \textbf{0.8576}$\pm$\textbf{0.01} & 0.8246$\pm$0.01 & \underline{0.8557$\pm$0.03} & 0.8464$\pm$0.02 & 0.8326$\pm$0.01 & 0.8365$\pm$0.00 \\
    Human Identification & \underline{0.9987$\pm$0.00} & 0.9960$\pm$0.00 & 0.9896$\pm$0.01 & 0.9895$\pm$0.01 & 0.9977$\pm$0.00 & 0.9982$\pm$0.00 &  \textbf{0.9989}$\pm$\textbf{0.00} \\
    \midrule
    \textit{Average} & 0.8231$\pm$0.03 & 0.8301$\pm$0.02 & 0.8936$\pm$0.03 & 0.8547$\pm$0.05 & \underline{0.8990$\pm$0.04} & 0.8596$\pm$0.03 & \textbf{0.9082}$\pm$\textbf{0.03} \\ 
    \midrule
    
    \multicolumn{8}{l}{\textbf{Regression - MAE ($\downarrow$)}} \\
    \midrule
    Respiratory Rate & 1.628$\pm$0.05 & 1.415$\pm$0.11 & 1.261$\pm$0.02 & 3.323$\pm$0.11 & 1.069$\pm$0.03 & \underline{0.7289$\pm$0.03} & \textbf{0.2930}$\pm$\textbf{0.03} \\
    Heart Rate       & 1.396$\pm$0.09 & 1.086$\pm$0.07 & 1.268$\pm$0.11 & 2.773$\pm$0.07 & 1.005$\pm0.06$ & \underline{0.6558$\pm$0.10} & \textbf{0.4590}$\pm$\textbf{0.09}\\
    SpO2             & 2.569$\pm0.18$  & 1.784$\pm0.06$  & 1.265$\pm0.12$  & 4.052$\pm0.27$  & 1.902$\pm0.15$  & \underline{1.004$\pm0.18$}  & \textbf{0.1457}$\pm$\textbf{0.12} \\
    Systolic Blood Pressure      & 15.30$\pm$1.84 & 13.96$\pm$1.58 & \textbf{12.33}$\pm$\textbf{1.59} & 13.09$\pm$1.37 & 15.30$\pm$1.59 & 13.36$\pm$1.20  & \underline{12.74$\pm$1.48} \\
    Diastolic Blood Pressure     & 8.557$\pm$0.93  & 8.944$\pm$0.99 & 8.695$\pm$0.75  & 9.211$\pm$0.78 & \underline{8.477$\pm$0.62} & 8.486$\pm$0.59  & \textbf{8.428}$\pm$\textbf{0.74} \\
    Average Heart Rate       & 4.354$\pm0.56$  & 4.174$\pm0.53$  & \textbf{3.697}$\pm$\textbf{0.66}  & 4.135$\pm0.53$ & 4.337$\pm0.33$ & 3.967$\pm$0.68  & \underline{3.805$\pm$0.57} \\
    
    \midrule
    \textit{Average} & 5.634$\pm$0.61 & 5.227$\pm$0.56 &4.753$\pm$0.54 & 6.097$\pm$0.52 & 5.348$\pm$0.46 &  \underline{4.700$\pm$0.46} & \textbf{4.312}$\pm$\textbf{0.51} \\ 
    \bottomrule
    \end{tabular}
}
\end{table*}

\begin{table*}[t]
\centering
\caption{\textbf{Linear Probing Performance.} Comparison of our SIGMA-PPG models against state-of-the-art baselines under the linear probing setting. Best results are \textbf{bold}, and second best are \underline{underlined}.}
\label{tab:linear_probing_results}

\resizebox{\linewidth}{!}{
    \setlength{\tabcolsep}{4pt} 
    \begin{tabular}{l|lllll|ll}
    \toprule
    \multirow{2}{*}{\textbf{Classification - AUC ($\uparrow$)}} 
    & \textbf{PAPAGEI-S} \scriptsize{(5M)} & \textbf{PAPAGEI-P} \scriptsize{(5M)} & \textbf{Pulse-PPG} \scriptsize{(28M)} 
    & \textbf{AnyPPG}\scriptsize{(5M)} & \textbf{GPT-PPG}\scriptsize{(19M)}
    & \textbf{SIGMA-PPG} \scriptsize{(5M)} & \textbf{SIGMA-PPG} \scriptsize{(30M)} \\
    
     & \citep{pillai2024papagei} & \citep{pillai2024papagei} & \citep{saha2025pulse}
     & \citep{nie2025anyppg} & \citep{chen2025gpt}
     & \multicolumn{1}{c}{(Ours)} & \multicolumn{1}{c}{(Ours)} \\
    \midrule

    Stress       & 0.3974$\pm$0.01  & 0.7880$\pm$0.02  & \textbf{0.9687}$\pm$\textbf{0.07} & \underline{0.9589$\pm$0.04} & 0.7767$\pm$0.08 &  0.8204$\pm$0.05 & 0.9083$\pm$0.02  \\
    Affect       & 0.7684$\pm$0.05 & 0.5059$\pm$0.01  & \underline{0.7807$\pm$0.03}  & \textbf{0.8288}$\pm$\textbf{0.02} & 0.6186$\pm$0.05 & 0.6822$\pm$0.04 & 0.7702$\pm$0.02 \\
    Hypertension & 0.7201$\pm0.17$  & 0.5253$\pm0.12$  & \textbf{0.7971$\pm$0.15}  & 0.4474$\pm0.07$  & 0.6467$\pm0.10$ & 0.6478$\pm0.08$  & \underline{0.7655$\pm0.08$} \\
    Signal Quality & 0.6446$\pm$0.03 & 0.6504$\pm$0.03 & \underline{0.9514$\pm$0.01}  & 0.7345$\pm$0.02 & 0.9439$\pm$0.02 & 0.9484$\pm$0.03  &  \textbf{0.9601}$\pm$\textbf{0.01} \\
    Activity & 0.7043$\pm$0.01 & 0.7163$\pm$0.02 & \underline{0.8051$\pm$0.01} & 0.7657$\pm$0.04 & 0.7576$\pm$0.02 & 0.7637$\pm$0.01 &  \textbf{0.8114}$\pm$\textbf{0.03} \\
    Human Identification & 0.7523$\pm$0.01 & 0.8731$\pm$0.01 & \textbf{0.9892}$\pm$\textbf{0.02} & \underline{0.9704$\pm$0.00} & 0.8316$\pm$0.02 & 0.8894$\pm$0.01 &  0.8906$\pm$0.01 \\
    \midrule
    \textit{Average} & 0.6645$\pm$0.05 & 0.6765$\pm$0.04 & \textbf{0.8820}$\pm$\textbf{0.05} & 0.7960$\pm$0.03 & 0.7625$\pm$0.05 & 0.7920$\pm$0.04 & \underline{0.8510$\pm$0.03} \\ 
    \midrule
    
    \multicolumn{8}{l}{\textbf{Regression - MAE ($\downarrow$)}} \\
    \midrule
    Respiratory Rate         & 3.632$\pm0.10$ & 2.410$\pm0.02$ & 2.054$\pm0.07$ & 4.034$\pm0.08$ & 2.018$\pm$0.03  & \underline{1.904$\pm0.02$} & \textbf{1.880}$\pm$\textbf{0.03} \\
    Heart Rate               & 22.56$\pm0.49$ & 10.53$\pm0.22$ & 6.946$\pm0.26$ & 19.0$\pm0.16$ & 9.273$\pm0.34$   & \underline{6.171$\pm0.14$} &  \textbf{5.831}$\pm$\textbf{0.19} \\
    SpO2                     & 24.05$\pm0.51$ & 6.558$\pm0.32$ & 5.185$\pm0.18$ & 25.14$\pm0.17$ & \textbf{3.174}$\pm$\textbf{0.13} & 5.114$\pm0.09$ & \underline{4.557$\pm0.19$} \\
    Systolic Blood Pressure  & 21.65$\pm$1.83 & 24.79$\pm$1.68 & \underline{13.62$\pm$0.97}  & 18.05$\pm$0.83 & 15.84$\pm$1.20 & 15.13$\pm$1.08 & \textbf{13.07}$\pm$\textbf{1.03} \\
    Diastolic Blood Pressure & 9.246$\pm$2.60 & 9.835$\pm$0.53 & 8.878$\pm$0.62 & 11.60$\pm$0.44 & \underline{8.585$\pm$0.99} & 9.321$\pm$0.93 & \textbf{7.748$\pm$0.80} \\
    Average Heart Rate       & 8.421$\pm2.44$ & 8.568$\pm2.31$ & \textbf{4.003}$\pm$\textbf{0.22}  & 11.31$\pm1.63$ & 8.897$\pm0.87$ & 7.342$\pm1.33$ & \underline{6.729$\pm0.27$}\\
    \midrule
    \textit{Average}         & 14.93$\pm$1.33 & 10.45$\pm$0.85 & \underline{6.781$\pm$0.38} & 14.86$\pm$0.55 & 7.965$\pm$0.59 & 7.497$\pm$0.60 & \textbf{6.636}$\pm$\textbf{0.42} \\ 
    \bottomrule
    \end{tabular}
}
\end{table*}

\subsection{Comparison with the state-of-the-art models}

The detailed information of the models considered as baseline can be found in the Appendix~\ref{sec:baseline}. First, we validated the necessity of pre-training compared to learning from scratch, i.e. random initialization,  in Appendix \ref{app:pretraining}. We've comprehensively evaluated SIGMA-PPG against promising baselines across two evaluation protocols: Full Fine-tuning and Linear Probing. The results are summarized in Table~\ref{tab:main_results_clean} and Table~\ref{tab:linear_probing_results}. For comprehensive results including additional metrics, please refer to Appendix~\ref{sec:sota}. All state-of-the-art (SOTA) methods were initialized with official weights. We further compared the promising Pulse-PPG algorithm by retraining it on the same clinical dataset used for SIGMA-PPG, these ablation experiments are reported in Appendix~\ref{sec:pretraining data}.

\textbf{Full Fine-Tuning Performance.} 
Based on the scaling behavior analysis in Appendix \ref{sec:model scale}, we selected the 30M parameter model as an optimal trade-off. As shown in Table~\ref{tab:main_results_clean}, SIGMA-PPG (30M) achieves SOTA performance across most tasks. Notably, in regression tasks involving precise scalar estimation, our model shows a clear advantage. For instance, in SpO2 estimation, SIGMA-PPG achieves a Mean Absolute Error (MAE) of 0.1457, representing an order-of-magnitude improvement over the best contrastive baseline (Pulse-PPG MAE = 1.265). Similarly, for respiratory rate and blood pressure estimation, our model consistently outperforms baselines.

We attribute this superiority to the generative masked modeling paradigm. When all parameters are unfrozen during fine-tuning, this morphology-aware initialization enables rapid adaptation, capturing precise waveform dynamics required for high-precision vital sign estimation and classification.

\textbf{Linear probing analysis.} 
In the linear probing setting (Table~\ref{tab:linear_probing_results}), where the pre-trained encoder is frozen, SIGMA-PPG further exhibits robust performance, particularly in regression tasks. This indicates that our Prior-Guided pre-training successfully encodes intrinsic physiological parameters (e.g., for the blood pressure estimation) into the latent space without needing extensive adaptation. 
However, we observe a different performance in specific classification tasks, particularly the classification of stress condition and detection of emotions for the WESAD dataset and the identification of subjects on Real-World PPG, where Pulse-PPG remains the best model. This result can be attributed to two key factors:

\textit{Domain shift (clinical vs. ambulatory PPGs).} SIGMA-PPG is primarily pre-trained on clinical data from ICUs and operating rooms characterized by low rate of movements and similar devices (e.g. finger cuff devices). In contrast, the WESAD dataset is collected using a wrist-worn wearable device (Empatica E4) during intense stress protocols \citep{schmidt2018introducing}, introducing significant motion artifacts and morphological deviations inherent to the wrist vascular bed. Although Pulse-PPG was exposed to large-scale ambulatory data during pre-training, the frozen features of our model face a distribution shift when applied directly to noisy wearable data without fine-tuning.

\textit{Task alignment (generative vs. contrastive).} For human identification, contrastive objectives are naturally aligned with the task, as they explicitly maximize the separability between different subject instances. In contrast, our generative objective focuses on learning the universal underlying manifold of the PPG waveform. While this yields superior physiological understanding (hence the strong regression results), the frozen features may prioritize common hemodynamic structures over subject-specific identity markers.
Nevertheless, it is crucial to note that once fine-tuned (Table~\ref{tab:main_results_clean}), SIGMA-PPG bridges this gap effectively, e.g., achieving 0.9994 AUC for the classification of stress condition, demonstrating that our model learns a highly adaptable representation that can be easily realigned to ambulatory domains given minimal supervision.

\begin{figure*}[!t]
    \centering
    \includegraphics[width=\linewidth]{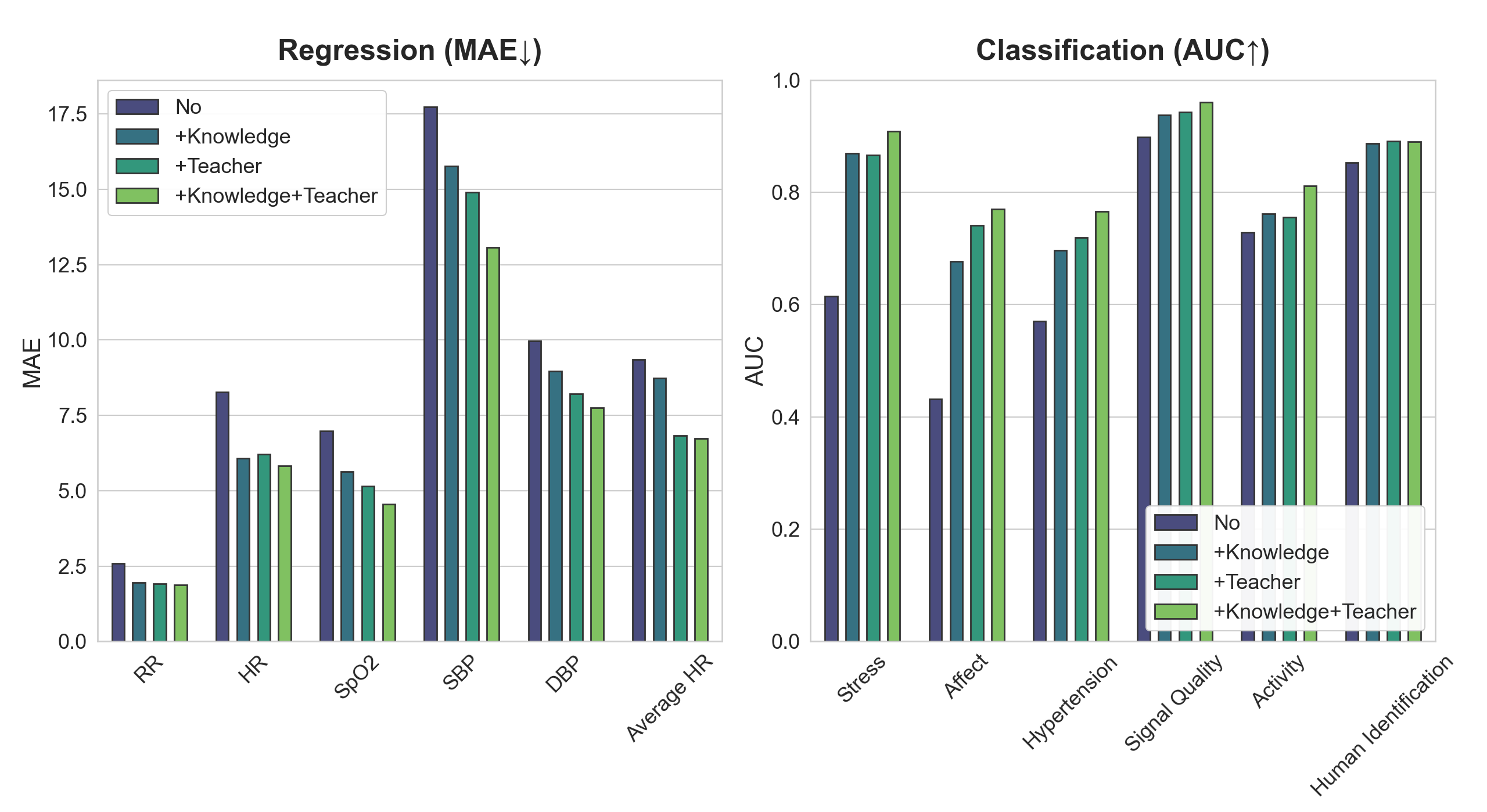}
    \caption{\textbf{Comparisons among different masking methods.} The colors of the bars correspond to four distinct masking mechanisms. \textbf{No.}: random masking, i.e. a standard uniform random masking without any prior knowledge; \textbf{+knowledge}: static prior-based probabilistic masking, i.e. a heuristic approach where masking probability is statically determined by skewness and amplitude; \textbf{+Teacher}: unconstrained adversarial masking, i.e. dynamic masking generated by a Teacher without prior constraints; \textbf{+knowledge+Teacher}: the proposed Prior-Guided Adversarial Masking, where the Teacher efficiently targets physiologically significant structures under the guidance of statistical priors.}
    \label{fig:mask_ablation}
\end{figure*}

\subsection{Effectiveness of Prior-Guided Adversarial Masking}
To validate the efficacy of our proposed Prior-Guided Adversarial Masking strategy, we've compared the behavioral patterns of different masking policies during the pre-training phase: random masking, static prior-based probabilistic masking, unconstrained adversarial masking, i.e. Teacher without Prior Bias, and the proposed Statistical-Prior Informed Adversarial Masking, i.e. Teacher with Prior Bias. Our method resulted superior in both regression and classification tasks.

\textbf{Irrelevance of random masking.} As illustrated in Figure \ref{fig:mask_ablation}, random masking creates uniformly distributed gaps along the time axis. Given the high morphological redundancy and quasi-periodicity of PPG signals, this strategy results overly simplistic. The Student model can recover missing values via simple local interpolation from adjacent time points without comprehending the global semantic structure of the waveform. Consequently, the model converges to a trivial solution, failing to extract robust physiological representations. 

\textbf{Limitations of static prior-based masking.} We evaluate a deterministic heuristic that sets masking probabilities from statistical priors (amplitude and skewness). Although it targets high-information regions better than random masking, its static nature risks overfitting salient areas while neglecting informative transitional segments.

\textbf{Degenerate solution in unconstrained adversarial training.} Introducing an adversarial Teacher without prior constraints leads to a degenerate solution: to maximize reconstruction error, the Teacher targets noisy or artifact-dominated PPG segments, causing the Student to overfit irrelevant patterns instead of learning meaningful physiological structure.

\textbf{Adjustment via statistical priors.} Incorporating physiological statistical priors into the Teacher’s reward reshapes its optimization, steering masking away from noise and toward physiologically salient regions, particularly systolic peaks. This forces the Student to learn underlying periodicity and morphology rather than relying on interpolation, yielding representations with strong physiological semantic consistency through prior-guided curriculum learning. Examples of how different masking strategies operate, can be found in Appendix~\ref{sec:masking}.

\section{Conclusion}

In this paper, we presented SIGMA-PPG, a masked reconstruction framework designed to mitigate the inherent redundancy and noise in physiological signal learning. The proposed approach integrates Prior-Guided Adversarial Masking with a Vector Quantized Variational Autoencoder (VQ-VAE) architecture. A key component of the framework is a semantic consistency loss imposed within the quantized latent space, which encourages morphologically similar waveforms, despite potential perturbations, to be mapped into the same codebook indices. This effectively disentangles physiological semantics from nuisance variability. 
Experimental results demonstrate that this constraint enables the model to capture fine-grained hemodynamic features more effectively than contrastive baselines, particularly in case of regression tasks. Although our findings indicate that fine-tuning is still necessary for robust cross-domain generalization, the proposed architecture provides a systematic methodological framework for leveraging discrete representations and adversarial priors. While future work is needed to address broader device variability, this study offers a solid step forward in the development of robust biomedical signal foundation models.

\section*{Impact statement}

This work aims to advance machine learning research by developing a foundation model for physiological signal understanding. By improving the robustness and generalizability of PPG representations, SIGMA-PPG has the potential to support more reliable cardiovascular monitoring, disease risk assessment, and long-term health tracking across both clinical and wearable settings. We anticipate that these methodological advances may contribute to safer, more accessible, and data-efficient health technologies, enabling earlier detection of physiological changes and supporting personalized and preventive healthcare.


\bibliography{main}
\bibliographystyle{plainnat}

\newpage
\appendix
\onecolumn
\section*{Appendix}
\section{Pre-training data and settings}
\label{sec:pretraining settings}

\textbf{Pre-training data.} We used two datasets. \textit{VitalDB} contains data collected from patients undergoing routine or emergency surgery in 10 of the 31 operating rooms at Seoul National University Hospital in South Korea. It includes a total of 6,388 surgical patients, encompassing 557,622 distinct data tracks. Signals are sampled at 500Hz. \textit{MIMIC-III Waveform Database Matched Subset} is a collection of waveform data obtained from bedside patient monitors in both adult and neonatal ICU. It comprises a total of 22,317 waveform records from 10,282 distinct ICU patients, sampled at a frequency of 125 Hz. Importantly, since the BIDMC PPG and Respiration Dataset used for downstream tasks are derived from the MIMIC repository, careful measures were taken to prevent data leakage. To ensure a strict separation between pre-training and downstream evaluation, all subjects appearing in the MIMIC III Matched Waveform Database were explicitly identified and removed from the pre-training dataset.

\textbf{Signal pre-processing.} To assure high-quality single-channel PPG signals across all datasets, we perform the following pre-processing analyses: (1) We apply a zero-phase 2nd-order Butterworth bandpass filter with low and high pass cut-offs set at 0.5Hz and 8Hz, respectively, to eliminate baseline wander and high-frequency noise; (2) the filtered signals are segmented into non-overlapping 4-minute windows and resampled to a sampling rate of 50Hz; (3) we implement a rigorous artifact rejection protocol, discarding segments containing flatlines or exceeding a 20\% missing value (NaN), while applying linear interpolation to impute segments with less than 20\% missing data; and (4) finally, we normalize each valid segment to the [0, 1] range using Min-Max scaling to standardize signal amplitudes across different acquisition devices.

\textbf{Model architecture}
\label{architecture}
SIGMA-PPG architecture is designed to be flexible and scalable, offering multiple model configurations ranging from Base (5.8M parameters) to Huge (350M parameters). The model settings of tokenizer and pre-training models are shown in Table \ref{tabel:tokenizer model} and Table \ref{tabel:pretraining model}, respectively.

\begin{table}[!h]
\centering
\caption{\textbf{Hyperparameters for vector-quantized tokenizer.}}
\label{tabel:tokenizer model}
\scalebox{0.85}{%
\begin{tabular}{lrc}
\toprule
\multicolumn{2}{c}{\textbf{Hyperparameters}} & \textbf{Values} \\
\midrule
\multirow{5}{*}{Temporal Encoder} & Input channels & \{1,8,8\} \\
 & Output channels & \{8,8,8\} \\
 & Kernel size & \{15,3,3\} \\
 & Stride & \{8,1,1\} \\
 & Padding & \{7,1,1\} \\
\midrule
\multicolumn{2}{c}{Transformer encoder layers} & 12 \\
\multicolumn{2}{c}{Transformer decoder layers} & 3 \\
\multicolumn{2}{c}{Hidden size} & 200 \\
\multicolumn{2}{c}{MLP size} & 800 \\
\multicolumn{2}{c}{Attention head number} & 10 \\
\multicolumn{2}{c}{Codebook size} & $4096 \times 64$ \\
\midrule
\multicolumn{2}{c}{Batch size} & 4096 \\
\multicolumn{2}{c}{Peak learning rate} & 3e-4 \\
\multicolumn{2}{c}{Minimal learning rate} & 1e-6 \\
\multicolumn{2}{c}{Learning rate scheduler} & Cosine \\
\multicolumn{2}{c}{Optimizer} & AdamW \\
\multicolumn{2}{c}{Adam $\beta$} & (0.9, 0.99) \\
\multicolumn{2}{c}{Weight decay} & 1e-4 \\
\multicolumn{2}{c}{Total epochs} & 100 \\
\multicolumn{2}{c}{Warmup epochs} & 10 \\
\multicolumn{2}{c}{Data stride} & 200 \\
\bottomrule
\end{tabular}%
}
\end{table}

\begin{table}[!h]
\centering
\caption{\textbf{Hyperparameters for pre-training model.}}
\label{tabel:pretraining model}
\resizebox{\linewidth}{!}{%
\begin{tabular}{lrcccc}
\toprule
\multicolumn{2}{c}{\textbf{Hyperparameters}} & \textbf{SIGMA-PPG-Base} & \textbf{SIGMA-PPG-Pro} & \textbf{SIGMA-PPG-Large} & \textbf{SIGMA-PPG-Large} \\
\midrule
\multirow{5}{*}{Temporal Encoder} & Input channels & \{1,8,8\} & \{1,8,8\} & \{1,16,16\} & \{1,32,32\} \\
 & Output channels & \{8,8,8\} & \{12,12,12\} & \{16,16,16\} & \{32,32,32\} \\
 & Kernel size & \multicolumn{4}{c}{\{15,3,3\}} \\ 
 & Stride & \multicolumn{4}{c}{\{8,1,1\}} \\
 & Padding & \multicolumn{4}{c}{\{7,1,1\}} \\
\midrule
\multicolumn{2}{r}{Transformer encoder layers} & 12 & 18 & 24 & 48 \\
\multicolumn{2}{r}{Hidden size} & 200 & 360 & 400 & 800 \\
\multicolumn{2}{r}{MLP size} & 800 & 800 & 1600 & 3200 \\
\multicolumn{2}{r}{Attention head number} & 10 & 12 & 16 & 16 \\
\midrule
\multicolumn{2}{r}{Batch size} & \multicolumn{4}{c}{4096} \\
\multicolumn{2}{r}{Peak learning rate} & \multicolumn{4}{c}{3e-3} \\
\multicolumn{2}{r}{Minimal learning rate} & \multicolumn{4}{c}{1e-5} \\
\multicolumn{2}{r}{Learning rate scheduler} & \multicolumn{4}{c}{Cosine} \\
\multicolumn{2}{r}{Optimizer} & \multicolumn{4}{c}{AdamW} \\
\multicolumn{2}{r}{Adam $\beta$} & \multicolumn{4}{c}{(0.9,0.98)} \\
\multicolumn{2}{r}{Weight decay} & \multicolumn{4}{c}{0.05} \\
\multicolumn{2}{r}{Total epochs} & \multicolumn{4}{c}{80} \\
\multicolumn{2}{r}{Warmup epochs} & \multicolumn{4}{c}{10} \\
\multicolumn{2}{r}{Data stride} & \multicolumn{4}{c}{800} \\
\midrule
\multicolumn{2}{r}{Gradient clipping} & \multicolumn{4}{c}{2} \\
\multicolumn{2}{r}{Layer scale init} & 0.1 & 0.01 & 1e-5 & 1e-6 \\
\multicolumn{2}{r}{EMA weight} & \multicolumn{4}{c}{0.996} \\
\bottomrule
\end{tabular}%
}
\end{table}

\textbf{Model variants.} We design three distinct configurations of SIGMA-PPG: SIGMA-PPG-Base, SIGMA-PPG-Pro, SIGMA-PPG-Large and SIGMA-PPG-Huge. The parameter counts are approximately 5.8M for SIGMA-PPG-Base, 30M for SIGMA-PPG-Pro, 80M for SIGMA-PPG-Large, and 350M for SIGMA-PPG-Huge. In our experiments, we primarily focus on SIGMA-PPG-Pro (30M parameter model).

\textbf{Training implementation.} Our SIGMA-PPG foundation model was pre-trained on a high-performance computing cluster equipped with 16 NVIDIA A800 GPUs (80GB VRAM per card). The training regimen spanned 80 epochs with a global batch size of 4,096 and a learning rate set to $3 \times 10^{-3}$. Comprehensive hyperparameter configurations are detailed in Tables~\ref{tabel:tokenizer model} and~\ref{tabel:pretraining model}. We determined the optimal stopping point based on a dual convergence criterion: (i) the stabilization of the pre-training reconstruction loss, and (ii) the saturation of prediction accuracy for critical morphological landmarks, specifically the reconstruction of systolic peaks and flat baseline regions.

\textbf{Teacher architecture.} The Teacher network is designed as a lightweight Transformer Encoder to minimize computational overhead while capturing sufficient global context for mask generation.

\begin{itemize}
    \item \textbf{Input Representations:} The Teacher takes continuous raw PPG patches as input. These patches are projected into the latent space via a temporal convolutional patch embedding layer.
    
    \item \textbf{Network Depth \& Width:} The Teacher consists of only 2 layers of Transformer Encoder Blocks. The hidden dimension size is set to 64, with 4 attention heads per layer.
    
    \item \textbf{Parameter Count:} With these compact hyperparameters, the Teacher network contains approximately 0.1M parameters. This is significantly smaller than the Student network (e.g., 30M for SIGMA-PPG-Pro),
    
    \item \textbf{Output Head:} The output layer projects the hidden states to a scalar logit for each patch, resulting in an unnormalized logit sequence $L_{\text{Teacher}} \in \mathbb{R}^N$.
\end{itemize}

\section{Downstream Tasks}
\label{sec:downstream}

To comprehensively evaluate the effectiveness and generalization capability of our proposed method, we conducted experiments across a diverse set of downstream tasks, ranging from physiological parameter estimation to identification of subject and emotion classification. In the following sections, we describe the six datasets utilized for validation, covering both clinical and real-world ambulatory settings, and the specific experimental configurations and hyperparameter settings adopted for each task.

\subsection{Details of datasets}
\textbf{BIDMC PPG and Respiration Dataset \citep{pimentel2016toward}.} The signal were recorded from critically ill patients at the Beth Israel Deaconess Medical Center, this dataset comprises 53 recordings. Each recording has a duration of 8 minutes and includes photoplethysmogram (PPG), impedance respiratory signals, and electrocardiogram (ECG) signals, all sampled at 125 Hz. Physiological parameters such as heart rate (HR), respiratory rate (RR), and blood oxygen saturation (SpO2) are provided at a sampling rate of 1 Hz. To serve as the ground truth for validation, the dataset provides manual breath annotations performed by two independent annotators based on the impedance respiratory signal.

\textbf{PPG-BP Dataset \citep{liang2018new}.} This clinical dataset includes recordings from 219 participants, a significant portion of whom are diagnosed with hypertension. The data collection protocol involved a 10-minute relaxation period followed by the acquisition of three short finger PPG segments (2.1 seconds each) per participant. The dataset provides corresponding ground truth measurements for systolic blood pressure (BP), diastolic BP, and heart rate (HR), serving as a benchmark for regression and classification tasks.

\textbf{WESAD Dataset \citep{schmidt2018introducing}.} The dataset contains physiological signals from 15 subjects (12 males, 3 females) collected using a chest-based RespiBAN Professional and a wrist-worn Empatica E4. The E4 device records Blood Volume Pulse (BVP), Electrodermal Activity (EDA), and Skin Temperature (TEMP) at sampling rates of 64 Hz, 4 Hz, and 4 Hz, respectively. The study protocol consisted of five sessions: Baseline, Stress, Amusement, Meditation, and Rest. Stress was induced via the Trier Social Stress Test (TSST), comprising a public speaking task and a mental arithmetic task. Following standard evaluation protocols, we utilize the labeled data for both binary (stress vs. non-stress) and multi-class classification tasks.

\textbf{Stanford Dataset \citep{torres2020multi}.} A specialized dataset was constructed to provide ground truth labels for signal quality, derived from the study's larger collection of physiological photoplethysmography (PPG) signals. This dataset is obtained from the manual review of 1,000 randomly selected 25-second signal windows by experts. Based on established standardized criteria, such as Elgendi's quality index, each window was classified into one of three categories: 'Excellent' for high-fidelity signals with clear peaks and dicrotic notches, 'Acceptable' for signals with discernible peaks suitable for reliable heart rate estimation despite some noise, and 'Poor' for signals dominated by artifacts.

\textbf{PPG-DaLiA Dataset \citep{reiss2019deep}.} To evaluate physical activity classification in realistic settings, we utilize the PPG-DaLiA dataset. This multimodal dataset comprises physiological and motion data collected from 15 participants (7 males and 8 females, aged 21–55). Data acquisition was performed using a chest-worn RespiBAN Professional and a wrist-worn Empatica E4 device. The protocol simulated real-life scenarios with a total duration of approximately 2.5 hours per participant, encompassing eight distinct activities: sitting still, ascending/descending stairs, playing table soccer, cycling, driving, taking lunch breaks, walking, and working. Transition periods between these activities were labeled as a separate 'zero' class. Ground truth labels are provided for activity classes and heart rate.

\textbf{Real-World PPG Dataset \citep{siam2019real}.} The dataset consists of PPG recordings collected from 35 healthy subjects using a dedicated IoT sensor configuration. Each subject contributed between 50 and 60 individual signal segments. Each segment represents a 6-second recording sampled at 50 Hz, resulting in a sequence length of 300 data points per instance. For experimental validation, the dataset is partitioned into a training set containing 1,374 samples (approximately 66\%) and a testing set containing 700 samples (approximately 34\%).

\section{State-of-the-art models}
\label{sec:baseline}
Detailed characteristics of these baselines are summarized in Table \ref{tab:baseline_comparison}.

\begin{table}[!h]
  \centering
  \caption{\textbf{Characteristics of baseline PPG foundation models and SIGMA-PPG.} CL: Contrastive Learning; GPT: Generative Pre-trained Transformer; MAE: Masked Autoencoder; OR: Operating Room.}
  \label{tab:baseline_comparison}
    \begin{tabular}{lllll}
    \toprule
    \textbf{Model}  & \textbf{Domain} & \textbf{Sampling Rate} & \textbf{Window Size} & \textbf{Key Methodology} \\
    \midrule
    \textbf{PaPaGei-S/P}  & Clinical  & 125 Hz & 10s   & CL \\
    \textbf{Pulse-PPG} & Daily life & 50 Hz & 4min   & CL \\
    \textbf{AnyPPG}  & Clinical & 125 Hz & 10s & CL\\
    \textbf{GPT-PPG}  & Clinical & 40 Hz & 30s   & GPT \\
    \midrule
    \textbf{SIGMA-PPG}  & Clinical + OR & 50 Hz & 4min & MAE \\
    \bottomrule
    \end{tabular}
\end{table}

\subsection{Downstream tasks settings}
Table~\ref{baseline models-settings} provides a summary of the experimental settings and hyperparameter configurations used for evaluating the model on various downstream tasks. The evaluations cover three task types: Regression (R), binary classification (B), and multi-class classification (M). For each task, specific hyperparameters including the number of epochs, batch sizes, learning rates, and warmup epochs are detailed to ensure reproducibility and optimal performance.

\begin{table}[!h]
    \centering
    \caption{\textbf{Summary of experimental settings and hyperparameter configurations for the downstream tasks.} Task types are denoted as R (regression), B (binary classification), and M (multi-class classification). LR means leaning rate, and LoSo means leave-one-subject-out validation.}\label{baseline models-settings}
    
    \resizebox{\columnwidth}{!}{%
        \begin{tabular}{@{}llllllc@{}}
        \toprule
        \textbf{Dataset} & \textbf{Task (Type)} & \textbf{Epochs} & \textbf{Batch Size} & \textbf{LR} & \textbf{Warmup Epochs} & \textbf{Evaluation Method} \\ \midrule
        BIDMC & Respiratory Rate (R) & 100 & 64 &5e-4 & 20 & 5-fold CV  \\
              & Heart Rate (R) & 100 & 64 &5e-4 & 20 & 5-fold CV \\
              & SpO2 (R) & 100 & 64 &5e-4 & 20 & 5-fold CV \\
        PPG-BP & Systolic Blood Pressure (R) &100 & 32 &1e-3 & 20 & 5-fold CV \\
               & Diastolic Blood Pressure (R) &100 & 32 &1e-3 & 20 & 5-fold CV \\
               & Average Heart Rate (R) &100 & 32 &1e-3 & 20 & 5-fold CV \\
               & Hypertension (B) &100 & 32 &1e-3 & 20 & 5-fold CV  \\
        WESAD & Stress (B)& 50 & 32 &8e-5 & 10 & LoSo\\
              & Emotion (B)& 100 & 32 &5e-5 & 20 & LoSo\\ 
        Stanford & Signal Quality(M) & 100 & 2048 &5e-4 & 10 & 5-fold CV  \\
        DaLiA  & Activity(M) & 50 & 64 &1e-4 & 10 & LoSo  \\
        Real-World PPG & Subject Identification(M) & 100 & 64 &5e-4 & 20 & 5-fold CV  \\
              
              \bottomrule
        \end{tabular}%
    }
\end{table}

\section{Examples of different masking strategies}
\label{sec:masking}
\begin{figure}[!h]
    \centering
    \includegraphics[width=\linewidth]{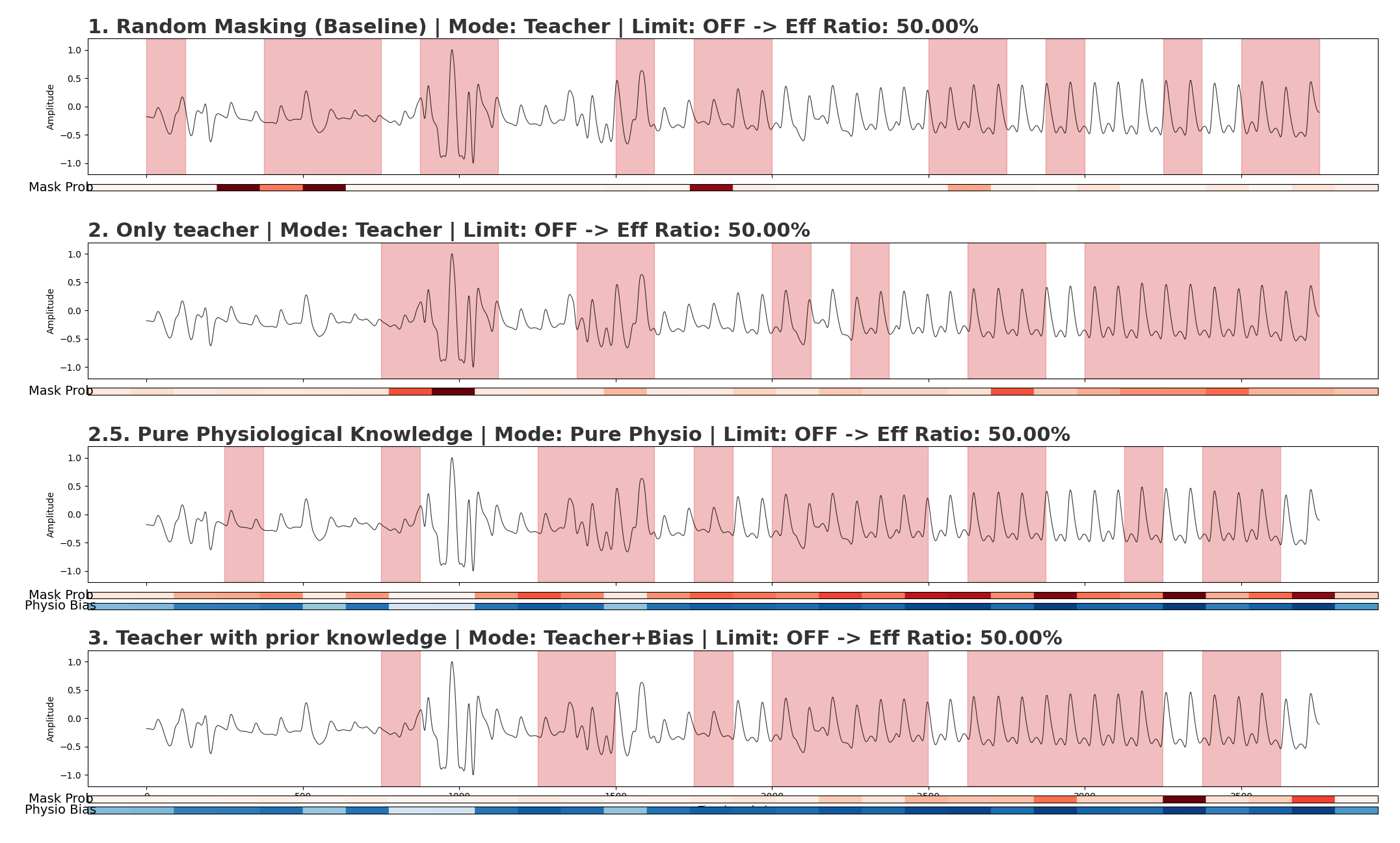}
    \caption{\textbf{Examples of different masking policies.} From the top panel to the bottom: Random Masking, Unconstrained Adversarial Masking, Static Prior Masking, and the proposed Prior-Guided Adversarial Masking. Only a 30-second signal segment is displayed.}
    \label{fig:adios}
\end{figure}

Figure~\ref{fig:adios} illustrates the results of different masking strategies.

\begin{itemize}
    \item \textbf{Random Masking:} this strategy samples indices uniformly without semantic awareness. Such segments are easily recoverable via simple local interpolation, rendering the pre-training objective trivial and failing to capture deep physiological semantics.
    
    \item \textbf{Unconstrained Adversarial (Teacher only):} this strategy maximizes the Student's reconstruction error, the Teacher exploits mathematically unpredictable stochastic noise. This solution forces the Student to memorize meaningless artifacts rather than learning robust morphological structures.
    
    \item \textbf{Static Prior-based:} utilizing skewness and amplitude stability constraints prevents the model from noise overfitting by prioritizing high-quality segments. However, this static approach lacks an adaptive curriculum, resulting in masks that are relatively easy to reconstruct and limiting the model's capacity to uncover global hemodynamic variations.
    
    \item \textbf{Prior-Guided Adversarial (SIGMA-PPG):} the proposed framework uses statistical priors to constrain the teacher to target patches that are  rich in physiological semantics. This approach generates challenging, spatially continuous masks that compel the Student to leverage global quasi-periodicity for reconstruction. To prevent training collapse, a span constraint limits the number of consecutively masked patches to a maximum of 5, balancing task difficulty with feasibility.
\end{itemize}

\section{Statistical-pior knowledge}
\label{sec:knowledge}
\begin{figure}[!h]
    \centering
    \includegraphics[width=\linewidth]{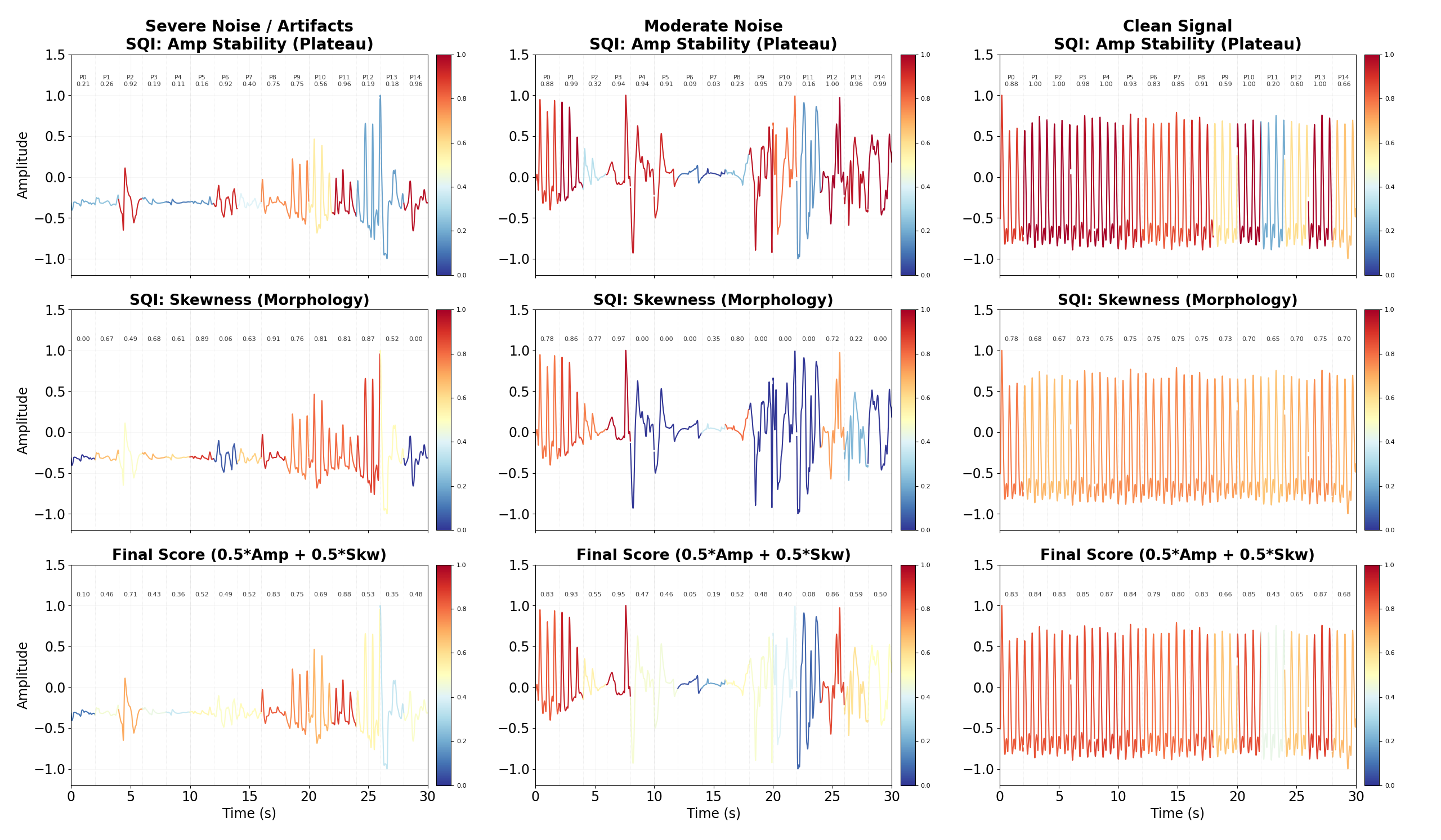}
    \caption{\textbf{Examples of how  statistical scoring functions are used.} Note how the score drops to near-zero during flat region ($t<20s$) but remains high ($\approx 1.0$) for clean, high-amplitude signals, effectively creating a ``Safe Zone'' for valid physiological data. For clarity's sake, just a 30-second signal segment is displayed. Colors provides indications of score values, from poor (blue) to high (red) quality signal.}
    \label{fig:prior_viz}
\end{figure}

Figure~\ref{fig:prior_viz} illustrates the dynamic behavior of the proposed statistical scoring function across three distinct levels of signal quality. It consists of three rows, each showing the raw PPG waveforms colored according to the corresponding amplitude stability score, skewness score, and final prior score.

\begin{itemize}
    \item \textbf{Scenario I: severe noise and artifacts.} The left panels illustrate a signal exhibiting non-physiological characteristics, flat lines could be due to possible signal loss and high-amplitude saturation to motion artifacts. In the flat region ($t < 20s$), the Amplitude Stability function $S_{amp}$ acts as a low-bound filter as it assumes values close to zero (blue line). Similarly, during the erratic high-frequency oscillations, the relative stability check based on the Median Absolute Deviation (MAD) $S_{rel}$ detects the distributional shift, keeping the final score $S_{prior}$  low. This ensures the Teacher network learns to mask these unrecoverable regions or ignore them, rather than forcing the Student to hallucinate features.

    \item \textbf{Scenario II: moderate noise.} The middle panels show a PPG signal containing baseline wander and some artifacts. While the signal is degraded, the quasi-periodic structure is partially preserved. Our scoring mechanism assigns an intermediate score ranging between 0.4 and 0.7, indicating to the model that these regions contain \textit{partial} information worth recovering, but with lower confidence than clean segments.

    \item \textbf{Scenario III: clean signal.} Right panels show a PPG waveform displaying clear systolic peaks and dicrotic notches. The \textit{trapezoidal flat-top window} used for the estimation of Amplitude Stability function $S_{amp}$, ensures that the high amplitude of these noisy-free beats does not trigger a penalty. Consequently, the final score $S_{prior}$ remains stable near 1.0, identifying this region as a high-value target for physiological feature extraction.
\end{itemize}

\section{Effectiveness of Semantic Consistency Constraint}
\label{sec:consistency loss}
\begin{figure*}[!h]
    \centering
    \includegraphics[width=\linewidth]{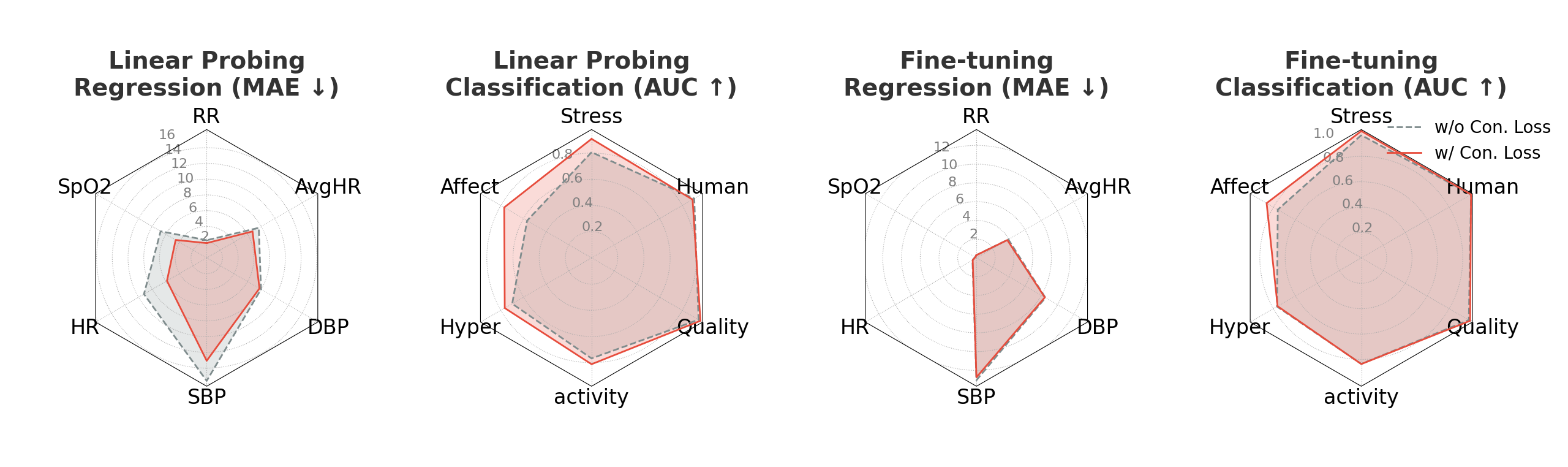}
    \caption{\textbf{Impact of semantic consistency constraint.} We compare the performance of SIGMA-PPG with (red) and without (grey) the consistency loss $\mathcal{L}_{Con}$ under Linear Probing  and Full Fine-tuning protocols, the two graphs on the left and on the right, respectively. The lower MAE and the higher AUC demonstrate that $\mathcal{L}_{Con}$ significantly enhances representation robustness against noise in regression tasks and improves semantic separability in classification tasks, serving as a critical component for both frozen feature extraction and downstream adaptation.}
    \label{fig:consis_loss_ablation}
\end{figure*}

We conducted a comprehensive ablation study by removing the semantic consistency constraint ($\mathcal{L}_{Con}$) from the pre-training phase while keeping all other components unchanged. We evaluated the model under both Linear Probing and Full Fine-tuning protocols. The comparative results are visualized in Figure~\ref{fig:consis_loss_ablation}.

\textbf{Robustness of frozen representations.} The impact of $\mathcal{L}_{Con}$ is most pronounced in the linear probing setting, which serves as a direct proxy for the intrinsic quality of the learned representations. As shown in the left panels of Figure~\ref{fig:consis_loss_ablation}, the model trained without consistency constraints (grey line) exhibits significant performance degradation. Specifically, in the regression tasks, the Mean Absolute Error (MAE) for Systolic Blood Pressure (SBP) deteriorates from 13.07 to 15.62, and for the heart rate (HR) error increases from 5.83 to 9.21. This indicates that without the explicit enforcement of consistency, the tokenizer is prone to overfitting to local morphological variations caused by motion artifacts or baseline wander, rather than capturing the underlying hemodynamic trends. By enforcing perturbed views of the same signal to map to identical codebook indices, $\mathcal{L}_{Con}$ effectively collapses the representation space, ensuring that nuisance variables are filtered out before quantization.

\textbf{Performance ceiling with fine-tuning.} When the model is full fine-tuned (Figure~\ref{fig:consis_loss_ablation}, right panels), i.e. when the encoder parameters are updated, the pre-trained weights informed by $\mathcal{L}_{Con}$ provide again a superior initialization point. We observe consistent improvements across all metrics, which confirms that the semantic alignment learned during pre-training is not transient, but provides a fundamental inductive bias that facilitates faster and better convergence on downstream tasks.

\section{Comparison with state-of-the-art models}
\label{sec:sota}

Table~\ref{tab:main_results_fine_tune} presents the performance of SIGMA-PPG model against SOTA baselines under the full fine-tuning setting. Specifically, we initialized all baselines using their official weights pre-trained on their respective datasets, followed by full parameter fine-tuning on the downstream tasks described in the Appendix~\ref{sec:baseline}.

\begin{table}[!h]
\centering
\caption{\textbf{Full Fine-Tuning Performances.} We compare the \textsc{Sigma-PPG} models against state-of-the-art baselines. The best results are highlighted in \textbf{bold}, and the second best results are \underline{underlined}.}
\label{tab:main_results_fine_tune}

\resizebox{\linewidth}{!}{%
    \setlength{\tabcolsep}{3.5pt} 
    \renewcommand{\arraystretch}{1.05} 
    \begin{tabular}{ll ccccc cc}
    \toprule
    \multicolumn{2}{c}{\multirow{2}{*}{\textbf{Task}}} & \multicolumn{5}{c}{\textbf{Baselines}} & \multicolumn{2}{c}{\textbf{Ours}} \\
    \cmidrule(lr){3-7} \cmidrule(lr){8-9}
    & & \textsc{Papagei-S} & \textsc{Papagei-P} & \textsc{Pulse-PPG} & \textsc{AnyPPG} & \textsc{GPT-PPG} & \textsc{Sigma-PPG} & \textsc{Sigma-PPG} \\
    & & \scriptsize{(5M)} & \scriptsize{(5M)} & \scriptsize{(28M)} & \scriptsize{(5M)} & \scriptsize{(19M)} & \scriptsize{(5M)} & \scriptsize{(30M)} \\
    \midrule
    
    \multicolumn{9}{l}{\textit{\textbf{Classification Tasks}}} \\
    \midrule
    \multirow{3}{*}{Stress} 
      & AUC ($\uparrow$) & 0.9601 & 0.9568 & 0.9852 & 0.9815 & \underline{0.9899} & 0.9256 & \textbf{0.9994} \\
      & Acc ($\uparrow$) & 0.8958 & 0.8820 & 0.9452 & 0.9583& \underline{0.9792} & 0.8854 & \textbf{0.9896} \\
      & F1 ($\uparrow$)  & 0.8214 & 0.7917 & 0.9392 & 0.9130 & \underline{0.9583} & 0.7925 & \textbf{0.9787} \\
    \cmidrule{1-9} 
    \multirow{3}{*}{Affect} 
      & AUC ($\uparrow$) & 0.7194 & 0.6945 & 0.7630 & 0.8407 & \underline{0.8516} & 0.7145 & \textbf{0.8620} \\
      & Acc ($\uparrow$) & 0.4792 & 0.5000 & \textbf{0.7079} & 0.6146 & 0.6042 & 0.4896 & \underline{0.6562} \\
      & F1 ($\uparrow$)  & 0.4542 & 0.4735 & 0.5582 & \underline{0.6506} & 0.6481 & 0.5147 & \textbf{0.6805} \\
    \cmidrule{1-9}
    \multirow{3}{*}{Hypertension} 
      & AUC ($\uparrow$) & 0.5149 & 0.5593 & \textbf{0.8258} & 0.5085 & 0.7496 & 0.7045 & \underline{0.7619} \\
      & Acc ($\uparrow$) & 0.9087 & \underline{0.9179 }& 0.7967 & 0.9148 & 0.9087 & 0.9132 & \textbf{0.9198} \\
      & F1 ($\uparrow$)  & 0.8655 & \textbf{0.8847} & 0.7213 & \underline{0.8809} & 0.8655 & 0.8690 & 0.8751 \\
    \cmidrule{1-9}
    \multirow{3}{*}{Signal Quality} 
      & AUC ($\uparrow$) & 0.8923 & 0.9162 & 0.9734 & 0.9523 & 0.9587 & \underline{0.9821} & \textbf{0.9907} \\
      & Acc ($\uparrow$) & 0.7433 & 0.7927 & 0.8921 & 0.8219 & 0.8502 & \underline{0.9093} & \textbf{0.9289} \\
      & F1 ($\uparrow$)  & 0.6524 & 0.6981 & 0.8279 & 0.7412 & 0.7651 & \underline{0.8414} & \textbf{0.8782} \\
    \cmidrule{1-9}
    \multirow{3}{*}{Activity} 
      & AUC ($\uparrow$) & 0.8533 & \textbf{0.8576} & 0.8246 & \underline{0.8557} & 0.8464 & 0.8326 & 0.8365 \\
      & Acc ($\uparrow$) & 0.3681 & 0.3754 & \textbf{0.4234} & \underline{0.4052} & 0.3681 & 0.3681 & 0.3643 \\
      & F1 ($\uparrow$)  & 0.3662 & \underline{0.3719} & 0.3648 & \textbf{0.3919} &  0.3579 &  0.3572 & 0.3432 \\
    \cmidrule{1-9}
    \multirow{3}{*}{Human Ident.} 
      & AUC ($\uparrow$) & \underline{0.9987} & 0.9960 & 0.9896 & 0.9895 & 0.9977 & 0.9982 &  \textbf{0.9989} \\
      & Acc ($\uparrow$) & 0.9253 & 0.8621 & 0.9571 & \underline{0.9590} & 0.9282 & 0.9474 & \textbf{0.9677} \\
      & F1 ($\uparrow$)  & 0.9167 & 0.8529 & 0.9570 & \underline{0.9592} & 0.9220 & 0.9435 & \textbf{0.9657}\\
    \midrule
    
    \multicolumn{9}{l}{\textit{\textbf{Regression Tasks}}} \\
    \midrule
    \multirow{3}{*}{Respiratory Rate} 
      & MAE ($\downarrow$) & 1.628 & 1.415 & 1.261 & 3.323 & 1.069 & \underline{0.7289} & \textbf{0.2930} \\
      & RMSE ($\downarrow$)& 2.344 & 1.965 & 1.716 & 3.976 & \underline{0.8498} & 1.101 & \textbf{0.4875} \\
      & Corr ($\uparrow$)  & 0.5306 & 0.7384 & 0.8092 & 0.2926 & 0.9545 & \underline{0.8974} & \textbf{0.9839} \\
    \cmidrule{1-9}
    \multirow{3}{*}{Heart Rate} 
      & MAE ($\downarrow$) & 1.396 & 1.086 & 1.268 & 2.773 & 1.005 & \underline{0.6558} & \textbf{0.4590} \\
      & RMSE ($\downarrow$)& 2.127 & 1.658 & 1.750 & 3.428 & 1.625 & \underline{1.0596} & \textbf{0.6771} \\
      & Corr ($\uparrow$)  & 0.9896 & 0.9930 & 0.9933 & 0.9883 & 0.9929 & \underline{0.9967} & \textbf{0.9987} \\
    \cmidrule{1-9}
    \multirow{3}{*}{SpO2} 
      & MAE ($\downarrow$) & 2.569 & 1.784 & 1.265 & 4.052 & 1.902 & \underline{1.004} & \textbf{0.1457} \\
      & RMSE ($\downarrow$)& 3.330 & 2.445 & \underline{1.616} & 4.790 & 3.099 & 1.767 & \textbf{0.2270} \\
      & Corr ($\uparrow$)  & 0.2821 & 0.7802 & \underline{0.9470} & 0.9123 & 0.3972 & 0.8194 & \textbf{0.9976} \\
    \cmidrule{1-9}
    \multirow{3}{*}{Systolic BP} 
      & MAE ($\downarrow$) & 15.30 & 13.96 & \textbf{12.33} & 13.09 & 15.30 & 13.36 & \underline{12.74} \\
      & RMSE ($\downarrow$)& 19.92 & 18.14 & \textbf{16.75} & \underline{16.98} & 19.58 & 19.66 & 18.23 \\
      & Corr ($\uparrow$)  & 0.2305 & 0.4329 & \textbf{0.5475} & \underline{0.5453} & 0.2850 & 0.2472 & 0.4828 \\
    \cmidrule{1-9}
    \multirow{3}{*}{Diastolic BP} 
      & MAE ($\downarrow$) & 8.557 & 8.944 & 8.695 & 9.211 & \underline{8.477} & 8.486 & \textbf{8.428} \\
      & RMSE ($\downarrow$)& 11.20 & 11.31 & 10.95 & 11.58 & 10.97 & \underline{10.93} & \textbf{10.86} \\
      & Corr ($\uparrow$)  & 0.6526 & 0.6023 & 0.6387 & 0.5993 & 0.6429 & \underline{0.6673} & \textbf{0.6758} \\
    \cmidrule{1-9}
    \multirow{3}{*}{Avg Heart Rate} 
      & MAE ($\downarrow$) & 4.354 & 4.174 & \textbf{3.697} & 4.135 & 4.337 & 3.967 & \underline{3.805} \\
      & RMSE ($\downarrow$)& 5.897 & 5.541 & \textbf{4.820} & 5.481 & 5.819 & 5.206 & \underline{5.149} \\
      & Corr ($\uparrow$)  & 0.8483 & 0.8705 & 0.8530 & 0.8659 & 0.8446 & \underline{0.8747} & \textbf{0.8748 }\\
    \bottomrule
    \end{tabular}%
}
\end{table}

\begin{table*}[!h]
\centering
\caption{\textbf{Linear Probing Performances.} Comparison of our SIGMA-PPG models against state-of-the-art baselines under the linear probing setting. The best results are highligthed in \textbf{bold}, and the second best results are \underline{underlined}.}
\label{tab:main_results_linear_prob}

\resizebox{\linewidth}{!}{%
    \setlength{\tabcolsep}{3.5pt} 
    \renewcommand{\arraystretch}{1.05} 
    \begin{tabular}{ll ccccc cc}
    \toprule
    \multicolumn{2}{c}{\multirow{2}{*}{\textbf{Task}}} & \multicolumn{5}{c}{\textbf{Baselines}} & \multicolumn{2}{c}{\textbf{Ours}} \\
    \cmidrule(lr){3-7} \cmidrule(lr){8-9}
    & & \textsc{Papagei-S} & \textsc{Papagei-P} & \textsc{Pulse-PPG} & \textsc{AnyPPG} & \textsc{GPT-PPG} & \textsc{Sigma-PPG} & \textsc{Sigma-PPG} \\
    & & \scriptsize{(5M)} & \scriptsize{(5M)} & \scriptsize{(28M)} & \scriptsize{(5M)} & \scriptsize{(19M)} & \scriptsize{(5M)} & \scriptsize{(30M)} \\
    \midrule
    
    \multicolumn{9}{l}{\textit{\textbf{Classification Tasks}}} \\
    \midrule
    \multirow{3}{*}{Stress} 
      & AUC ($\uparrow$) & 0.3974  & 0.7880  & \textbf{0.9687} & \underline{0.9589}  & 0.7767 &  0.8204 & 0.9083 \\
      & Acc ($\uparrow$) & 0.6849  & 0.7292  & \underline{0.8904 }& \textbf{0.9062}  & 0.6979 & 0.7604 &  0.8542 \\
      & F1 ($\uparrow$)  & 0.4065 & 0.6061 & \textbf{0.8759} & \underline{0.8000} & 0.5672 & 0.5660 & 0.7308 \\
    \cmidrule{1-9} 
    \multirow{3}{*}{Affect} 
      & AUC ($\uparrow$) & 0.7684 & 0.5059  & \underline{0.7807}  & \textbf{0.8288}  & 0.6186 & 0.6822 & 0.7702 \\
      & Acc ($\uparrow$) & 0.5169 & 0.2396  & \textbf{0.6517}  & \underline{0.6302}  & 0.3958 &  0.4029 & 0.5208 \\
      & F1 ($\uparrow$)  & 0.4010 & 0.2712 & 0.5431 & \textbf{0.5788} & 0.3956 & 0.4375  & \underline{0.5718} \\
    \cmidrule{1-9}
    \multirow{3}{*}{Hypertension} 
      & AUC ($\uparrow$) &  0.7201  & 0.5253  & \textbf{0.7971}  & 0.4474  & 0.6467 & 0.6533  & \underline{0.7655} \\
      & Acc ($\uparrow$) &  0.7398  & 0.8669  & 0.8130  & 0.8636  & 0.8636 & \underline{0.8864} &  \textbf{0.9087} \\
      & F1 ($\uparrow$)  & 0.6473 & 0.8013 & 0.6975 & 0.8004 & 0.8004 & \underline{0.8330} & \textbf{0.8655} \\
    \cmidrule{1-9}
    \multirow{3}{*}{Signal Quality} 
      & AUC ($\uparrow$) & 0.6446 & 0.6504 & \underline{0.9514} & 0.7345 & 0.9439 & 0.9484 &  \textbf{0.9601} \\
      & Acc ($\uparrow$) & 0.2151 & 0.2054 & \textbf{0.8195} & 0.2642 & 0.6950 & 0.6476 &  \underline{0.8075} \\
      & F1 ($\uparrow$)  & 0.1322 & 0.1230 & \underline{0.7256} & 0.1852 & 0.7234 & 0.7229 &  \textbf{0.7308} \\
    \cmidrule{1-9}
    \multirow{3}{*}{Activity} 
      & AUC ($\uparrow$) & 0.7043 & 0.7163 & \underline{0.8051} & 0.7657 & 0.7576 & 0.7637 &  \textbf{0.8114} \\
      & Acc ($\uparrow$) & 0.1981 & 0.2627 & \textbf{0.4104} & 0.3505 & 0.2264 & 0.2688 &  \underline{0.3733} \\
      & F1 ($\uparrow$)  & 0.1817 & 0.2464 & 0.3039 & \underline{0.3322} & 0.2093 & 0.2470 &  \textbf{0.3674} \\
    \cmidrule{1-9}
    \multirow{3}{*}{Human Ident.} 
      & AUC ($\uparrow$) & 0.7523 & 0.8731 & \textbf{0.9892} & \underline{0.9704} & 0.8316 & 0.8894 &  0.8906 \\
      & Acc ($\uparrow$) & 0.1152 & 0.2816 & \underline{0.7686} & 0.6037 & \textbf{0.8139} & 0.3399 & 0.3288 \\
      & F1 ($\uparrow$)  & 0.0762 & 0.2310 & \textbf{0.7476} & \underline{0.5401} & 0.5775 & 0.2634 & 0.2396 \\
    \midrule
    
    \multicolumn{9}{l}{\textit{\textbf{Regression Tasks}}} \\
    \midrule
    \multirow{3}{*}{Respiratory Rate} 
      & MAE ($\downarrow$)  & 3.632 & 2.410 & 2.054 & 4.034 & 2.018 & \underline{1.904} &  \textbf{1.880} \\
      & RMSE ($\downarrow$) & 4.859 & 3.062 & \underline{2.715} & 4.595 & 2.742 & 2.840 &  \textbf{2.662}  \\
      & Corr ($\uparrow$)   & 0.1265 & 0.1054  & \underline{0.3685}  & \textbf{0.3893}  & 0.2361 & 0.1829 & 0.3346  \\
    \cmidrule{1-9}
    \multirow{3}{*}{Heart Rate} 
      & MAE ($\downarrow$)  & 22.56 & 10.53 & 6.946 & 19.01 &  9.273 & \underline{6.171} &  \textbf{5.831} \\
      & RMSE ($\downarrow$) & 28.85 & 14.54 & \underline{9.092}  & 19.88 & 12.40 & 11.89 &  \textbf{8.139} \\
      & Corr ($\uparrow$)   & 0.2603 & 0.1305 & \underline{0.7339}  & 0.1921 & 0.5298 & 0.6366 & \textbf{0.7992}  \\
    \cmidrule{1-9}
    \multirow{3}{*}{SpO2} 
      & MAE ($\downarrow$)  & 24.05 & 6.558 & 5.185 & 25.14 & \textbf{3.174} & 5.114 &  \underline{4.557} \\
      & RMSE ($\downarrow$) & 29.83 & 8.863 & \underline{6.703} & 25.66 & \textbf{4.480} & 8.495 &  \underline{6.916} \\
      & Corr ($\uparrow$)   & 0.4677  & 0.8802  & \underline{0.9216}  & 0.5234  & \textbf{0.9621} & 0.8827  & 0.9163 \\
    \cmidrule{1-9}
    \multirow{3}{*}{Systolic BP} 
     & MAE ($\downarrow$)  & 21.65 & 24.79 & \underline{13.62} & 18.05 & 15.84 & 15.13 &  \textbf{13.07} \\
      & RMSE ($\downarrow$) & 23.87 & 30.59 & \underline{16.93} & 22.99 & 20.27 & 21.36 &  \textbf{16.01} \\
      & Corr ($\uparrow$)   & 0.3165 & 0.2485  & \underline{0.5259}  & 0.2906 & 0.1284 & 0.1565  &  \textbf{0.6029}\\
    \cmidrule{1-9}
    \multirow{3}{*}{Diastolic BP} 
      & MAE ($\downarrow$)  & 9.246 & 9.835 &  8.878 & 11.60 & \underline{8.585} & 9.321 & \textbf{7.748} \\
      & RMSE ($\downarrow$) & 11.27  & 12.64 & \underline{10.87}  & 14.28 & 11.17 & 11.87 &  \textbf{9.970} \\
      & Corr ($\uparrow$)   & 0.5401 & 0.5294  & \underline{0.5973}  & 0.1472  & 0.3477 & 0.1325  &  \textbf{0.6812} \\
    \cmidrule{1-9}
    \multirow{3}{*}{Avg Heart Rate} 
      & MAE ($\downarrow$)  & 8.421 & 8.568 & \textbf{4.003} & 11.31 & 8.897 & 7.342 &  \underline{6.729} \\
      & RMSE ($\downarrow$) & 12.33 & 11.06 & \textbf{5.200} & 14.62 & 10.91 & 9.193 &  \underline{8.5119} \\
      & Corr ($\uparrow$)   & 0.3109  & 0.3424  & \textbf{0.6870}  & 0.0934  & 0.3968 & 0.5308 &  \underline{0.5244} \\
    \bottomrule
    \end{tabular}%
}
\end{table*}

Table~\ref{tab:main_results_linear_prob} presents the comprehensive performance of SIGMA-PPG against other SOTA baselines when Linear Probing is used. Specifically, we initialized all baselines using their official weights pre-trained on their respective datasets, followed by training only the linear head of our downstream tasks while keeping the backbone parameters frozen.

\section{Geometric validation of semantic consistency constraint}
\label{app:consistency_validation}
\begin{figure}[h]
    \centering
    \includegraphics[width=1.0\textwidth]{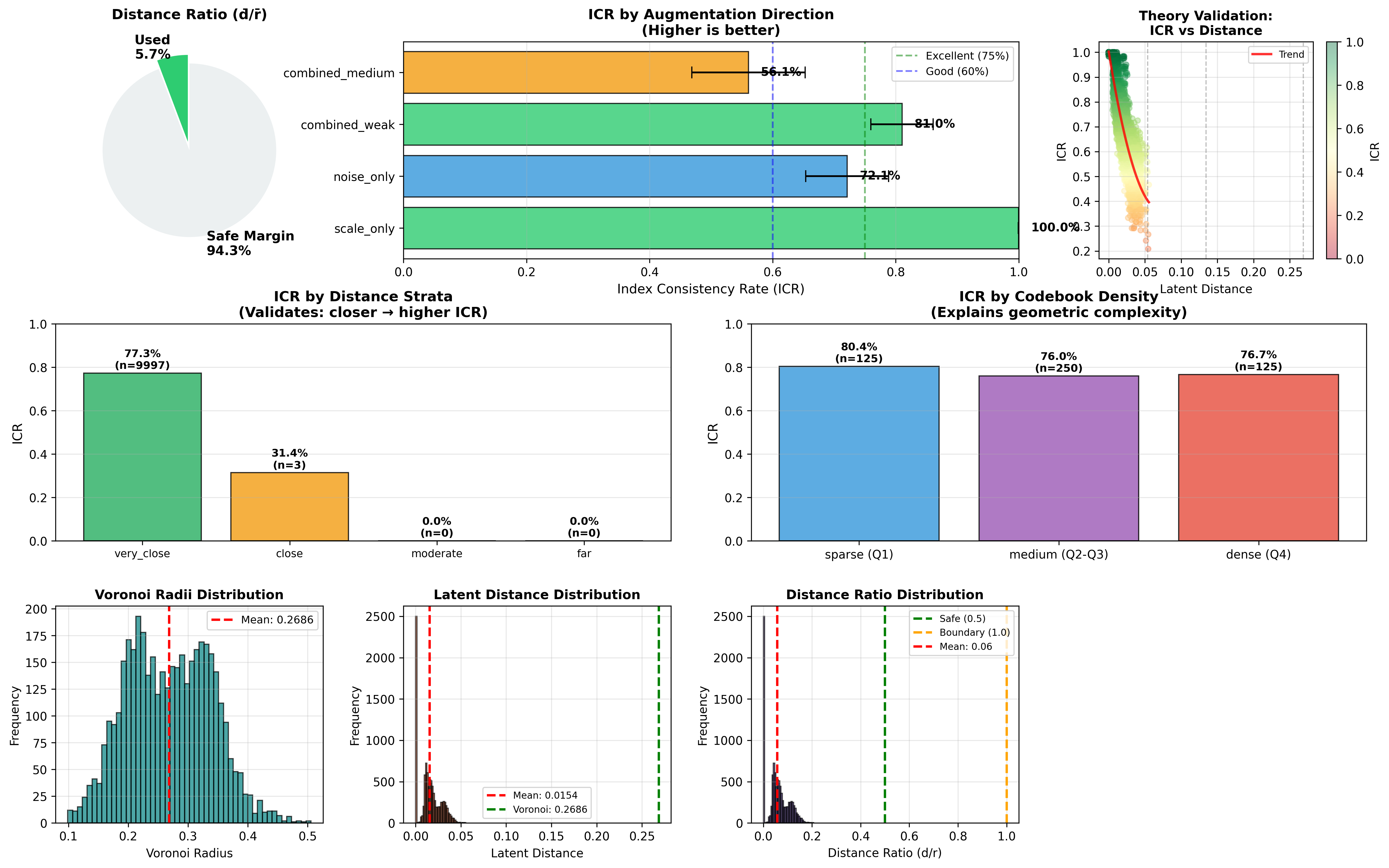}
    \caption{\textbf{Geometric validation of semantic consistency constraint.} First row: (Left) Distance Ratio ($d/r$) provides $5.9\%$ of Voronoi radius, i.e. $94.1\%$ of safety margin. (Middle) Index Consistency Rate (ICR) by augmentation: scale-only achieves $100\%$, training-matched $80.5\%$ (combined\_weak), noise sensitivity (noise\_only) and a moderate combination of scale and noise (combined\_medium) are shown as well. (Right) Monotonic negative correlation between distance and ICR validates theoretical framework. Second row: (Left) Distance-stratified ICR: $76.8\%$ for $d/r < 0.2$ ($99.95\%$ of samples). (Right) Density-stratified ICR: robust $75.8--79.2\%$ across quartiles. third row: (Left) Voronoi radii  distribution (mean=0.269), (Middle) Latent distances distribution (mean=0.016) , (Right) Distance ratios distribution (mean=0.06, $99.9\%$ below safety threshold 0.5).}
    \label{fig:consistency_validation}
\end{figure}

Geometric validation of the semantic consistency constraint assesses whether the imposed constraint induces a meaningful latent geometry, in which semantically similar signals are mapped to nearby representations or shared discrete tokens despite nuisance variations.
To test the effectiveness of the proposed semantic consistency constraint in the discrete codebook space, we performed a geometric validation on 500 held-out samples. While $\mathcal{L}_{\text{con}}$ operates on continuous pre-quantization features, our results demonstrate it provides a probabilistic guarantee of discrete consistency. Results are shown in Figure~\ref{fig:consistency_validation}.

\textbf{Distance ratio and Voronoi safety margin.}
Augmentation-induced latent shifts were quantified using the Distance Ratio $d/r $, which compares the embedding displacement caused by semantic-preserving augmentations to the average inter-sample distance, thereby assessing representation stability as:
\begin{equation}
    d/r = \frac{\|\mathbf{z}_e(x) - \mathbf{z}_e(x')\|_2}{\bar{r}_{\text{Voronoi}}}, \quad \bar{r}_{\text{Voronoi}} = \frac{1}{2}\min_{k \neq k'} \|\mathbf{e}_k - \mathbf{e}_{k'}\|_2
\end{equation}
where $\mathbf{z}_e(x)$ and $\mathbf{z}_e(x')$ are the latent representations of the original and augmented views.

In the first row, the left panel of Figure~\ref{fig:consistency_validation} shows a mean $d/r = 0.059$ (5.9\%). This Distance Ratio indicates that augmentation-induced embedding shifts are an order of magnitude smaller than inter-sample distances, resulting in highly compact semantic clusters and a large safety margin against representation collapse or semantic confusion as shown also in the lower panels.

\textbf{Direction-specific index consistency.}
The Index Consistency Rate (ICR) quantifies the proportion of cases in which SIGMA-PPG preserves the correct directional trends of PPG-derived indices under semantic-preserving perturbations and domain shifts.
ICR analysis across different augmentation types is illustrated in Figure~\ref{fig:consistency_validation}, first row, middle panel, from the bottom to the top bar:

\begin{itemize}
    \item \textit{Scale invariance:} amplitude scaling ($\pm 3\%$) achieves 100\% ICR, confirming perfect morphology-amplitude decoupling.
    
    \item \textit{Training-matched:} weak perturbations (scale $\pm 2\%$, noise $\sigma=0.02$) yield 80.5\% ICR ($\pm 5.3\%$), demonstrating robust consistency under realistic augmentation.
    
    \item \textit{Noise sensitivity:} pure noise ($\sigma=0.03$) shows 72.8\% ICR, lower than scaling. This reflects directional sensitivity of Voronoi geometry: isotropic noise is more likely to encounter narrow cell corridors.
    
    \item \textit{Moderate augmentation:} stronger perturbations (scale $\pm 5\%$, noise $\sigma=0.05$) result in 55.0\% ICR. This selective inconsistency demonstrates discriminative capacity to reject severe artifacts, preventing codebook collapse.
\end{itemize}

\textbf{Theoretical validation.}
Figure~\ref{fig:consistency_validation},first row, right panel shows a clear monotonic negative correlation ($\rho = -0.87$) between distance and ICR. This suggests that greater latent separation leads to lower direction-specific index consistency.

\begin{itemize}
    \item \textit{Proxy validity:} distance minimization effectively maximizes discrete consistency, validating our surrogate objective.
    
    \item \textit{Stratified analysis:} ICR decreases from 76.8\% ($d/r < 0.2$, 99.95\% of samples) to 47.7\% ($0.2 \leq d/r < 0.5$). Concentration in "very close" stratum validates constraint effectiveness as shown in the second row, left panel).
\end{itemize}

\textbf{Density-stratified robustness.}
Figure~\ref{fig:consistency_validation}, second row, right panel shows ICR analysis for a stratification of local codebook density.
\begin{itemize}
    \item \textit{Stable performance:} ICR remains 75.8--79.2\% across quartiles ($\Delta < 3.4\%$), demonstrating uniform handling of geometric complexity.
    \item \textit{Non-degenerate partitioning:} balanced performance rules out underutilization or extreme density imbalance.
\end{itemize}

\textbf{Distributional regularity.}
Figure~\ref{fig:consistency_validation}, third row shows histograms of Voronoi radii, Latent distances and Distance ratios which confirm absence of irregularities.
\begin{itemize}
    \item \textit{Voronoi radii:} unimodal distribution (mean=0.269, std=0.062) indicates spatial uniformity. No heavy tails or degenerate cells.
    
    \item \textit{Latent distances:} Tight clustering (mean=0.016) with exponential decay demonstrates global constraint effectiveness, not subset cherry-picking.
    
    \item \textit{Distance ratios:} heavy left-skew (skewness=2.31), with 99.9\% below safety threshold ($d/r < 0.5$). Boundary crossing is rare, occurring under extreme out-of-distribution perturbations.
\end{itemize}

\begin{table}[!h]
\centering
\caption{\textbf{Performance comparisons for four different window durations.} Best results are highlighted in \textbf{bold}.}
\label{tab:window_size}
\begin{small} 
\begin{sc}    
\begin{tabular}{l|cccccc|c}
\toprule
\multicolumn{8}{c}{\textbf{Classification Task (AUC $\uparrow$)}} \\
\midrule
Time & Stress & Affect & Hyper & Quality & Activity & Human & \textit{Average} \\
\midrule
240s & \textbf{0.9083} & 0.7702 & 0.7655 & \textbf{0.9601} & \textbf{0.8314} & \textbf{0.8906} & \textbf{0.8544} \\
120s & 0.8755 & 0.6882 & \textbf{0.7826} & 0.9570 & 0.8310 & 0.8794 & 0.8356 \\
60s  & 0.8845 & 0.7626 & 0.6764 & 0.9396 & 0.7676 & 0.8404 & 0.8119\\
30s  & 0.8839 & \textbf{0.7864} & 0.7468 & 0.9296 & 0.7523 & 0.8359 & 0.8225 \\
\midrule
\multicolumn{8}{c}{\textbf{Regression Task (MAE $\downarrow$)}} \\
\midrule
Time & RR & HR & SpO2 & SBP & DBP & avgHR & \textit{Average} \\
\midrule
240s & \textbf{1.880} & \textbf{5.831} & 4.557 & 13.07 & \textbf{7.748} & \textbf{6.729} & \textbf{6.636} \\
120s & 2.265 & 10.96 & 4.21  & 14.71 & 8.496 & 7.631 & 8.045 \\
60s  & 2.406 & 9.867 & \textbf{3.095} & \textbf{12.48} & 8.408 & 7.385 & 7.274\\
30s  & 2.328 & 10.36 & 3.879 & 14.08 & 8.477 & 7.185 & 7.718 \\

\bottomrule
\end{tabular}
\end{sc}
\end{small}
\end{table}

\section{Effectiveness of window size}
\label{sec:window}

In this section, we examine how input window size affects downstream performance in order to establish the most suitable temporal context for learning physiological representations. We evaluate the performance of SIGMA-PPG model for four different window durations using Linear Probing: 30, 60, 120, and 240 seconds. The results from these comparisons are summarized in Table \ref{tab:window_size}.

\section{Detailed analysis of pre-training effectiveness}
\label{app:pretraining}

We benchmark the SIGMA-PPG model against a backbone model trained from scratch with random initialization to isolate the performance gains attributed to the pre-training paradigm. Table \ref{tab:ablation_pretrain} presents the comparison metrics across 10 downstream tasks.

\begin{table}[!h]
\centering
\caption{\textbf{Effectiveness of Pre-training.} Comparison between training the model from scratch, i.e. with random initialization, and the  full fine tuned SIGMA-PPG model. We report MAE for regression tasks ($\downarrow$) and AUC for classification tasks ($\uparrow$). The Imp.($\Delta)$ indicates the relative performance improvement.}
\label{tab:ablation_pretrain}
{\scshape
\begin{tabular}{lcccc}
\toprule
\textbf{Task Type} & \textbf{Task} & \textbf{Random Init.} & \textbf{SIGMA-PPG} & \textbf{Imp. ($\Delta$)} \\
\midrule
\multirow{3}{*}{\textbf{Regression (MAE $\downarrow$)}} 
 & Resp. Rate & 0.3178 & \textbf{0.2930} & +7.8\% \\
 & Heart Rate & 0.5010 & \textbf{0.4590} & +8.4\% \\
 & SpO2 & 0.1886 & \textbf{0.1457} & +22.7\% \\
 & Systolic BP & 17.59 & \textbf{12.74} & \textbf{+27.6\%} \\
 & Diastolic BP & 9.015 & \textbf{8.428} & +6.5\% \\
 & Avg. HR & 4.147 & \textbf{3.805} & +8.2\% \\
\midrule
\multirow{4}{*}{\textbf{Classification (AUC $\uparrow$)}} 
 & Stress & 0.9672 & \textbf{0.9994} & +3.3\% \\
 & Affect & 0.7605 & \textbf{0.8620} & \textbf{+13.3\%} \\
 & Hypertension & 0.6316 & \textbf{0.7619} & \textbf{+20.6\%} \\
 & Signal Quality & 0.9788 & \textbf{0.9907} & +1.2\% \\
 & activity & 0.8273 & \textbf{0.8365} & +1.1\% \\
 & Human Ident. & 0.9605 & \textbf{0.9989} & +4.0\% \\
\bottomrule
\end{tabular}
}

\end{table}

These findings confirm that the 120,000 hours of unlabeled pre-training equip SIGMA-PPG with strong inductive biases, effectively mitigating the data scarcity bottleneck inherent in medical AI tasks.

We can observe that, in most the cases, an increase in the window duration results in enhanced performance. This could be explained by  the inherent advantage of the Transformer architecture in modeling long-range dependencies \citep{guo2023remote}. The increased duration of the signal window may facilitate the self-attention mechanism in attending to stable periodic patterns over a span of thousands of time steps. This process is theorized to result in the suppression of high-frequency noise, thereby ensuring more stable and accurate physiological estimations. \citep{chen2024actnet}.

\begin{figure}[!h]
    \centering
    \includegraphics[width=\linewidth]{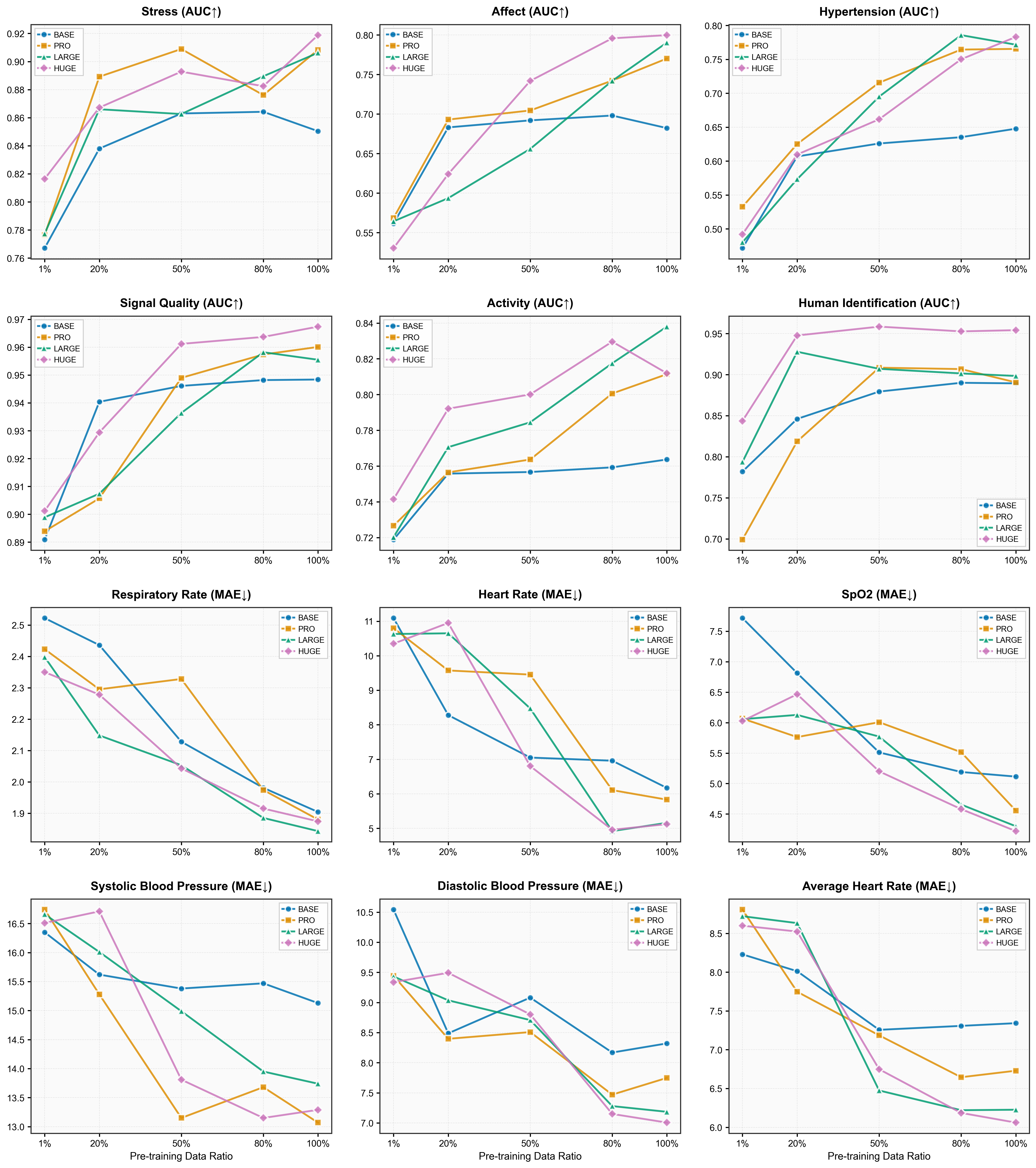}
    \caption{\textbf{Scaling behavior of SIGMA-PPG across different model sizes.} The panels illustrate the average Regression MAE (lower is better) and Classification AUC ( higher is better) for model variants with a number of parameters ranging from 5.8M (Base) to 350M (Huge). We observe consistent performance gains up to the Large model (80M), followed by a performance plateau or slight degradation at the Huge scale (350M).}
    \label{fig:scale}
\end{figure}

\section{Impact of model and data scale}
\label{sec:model scale}

To investigate the scalability of the SIGMA-PPG architecture and the interplay between model capacity and data volume, we provide a detailed analysis of the impact of model scale and data scale on downstream task performance.

\subsection{Experimental setup}
To ensure fair comparison and adherence to neural scaling laws, we strictly controlled experimental variables. We evaluated four model variants with different parameter counts: SIGMA-PPG-Base (5.8M), Pro (30M), Large (80M), and Huge (350M).
In order to eliminate the interference of improper optimization on the scalability analysis, we did not use fixed hyperparameters for all models. Instead, we strictly followed the scaling principles proposed by \citep{hoffmann2022training} to adjust the learning rate. Specifically, we set the peak learning rate ($\eta$) to be inversely proportional to the model parameter count $N$, following a power-law distribution:
$\eta \propto N^{-0.5}$.

This strategy ensures that all model variants, from Base to Huge,  are trained on optimization trajectories close to their theoretical optimum, thereby guaranteeing that performance differences stem primarily from model capacity itself rather than optimizer hyperparameter choices.

\subsection{Analysis of scaling behaviors}
Figure \ref{fig:scale} illustrates the average performance of models of different scales on downstream tasks. We observed the following two key phenomena:the positive scaling with model size and the anomaly of the Huge model.

Overall, the results demonstrate strong scalability of SIGMA-PPG model, with performance improving from Base (5.8M) to Pro (30M) and Large (80M) as model capacity increases. Error reductions are particularly pronounced for complex regression tasks such as SpO2 and blood pressure estimation, indicating that larger models better capture subtle PPG morphological features and yield more robust physiological representations.
However, further scaling the model to Huge (350M) does not yield additional gains and instead results in performance saturation or slight degradation compared to the Large (80M) model. This behavior can be interpreted in light of the Chinchilla scaling laws \citep{hoffmann2022training}. Moreover, despite using over 120,000 hours of pre-training data, this volume may be insufficient for a 350M-parameter Transformer to reach compute-optimal training. Given the higher redundancy and lower information density of PPG signals compared to language, the Huge model likely operates in an over-parameterized regime, leading to overfitting. Unlike discrete language data composed by symbols, physiological signals likely have a bounded intrinsic dimension. Beyond 80M parameters, additional model capacity may primarily fit non-stationary noise rather than meaningful structure, reducing downstream generalization.

In summary, although SIGMA-PPG exhibits strong scaling behavior, the 80M-parameter configuration (Large SIGMA-PPG) currently offers the optimal balance between performance and computational efficiency given the scale of available public data.

\section{Training loss}
\label{sec:training loss}

\subsection{Tokenizer pre-training}
\begin{figure}[!h]
    \centering
    \includegraphics[width=\linewidth]{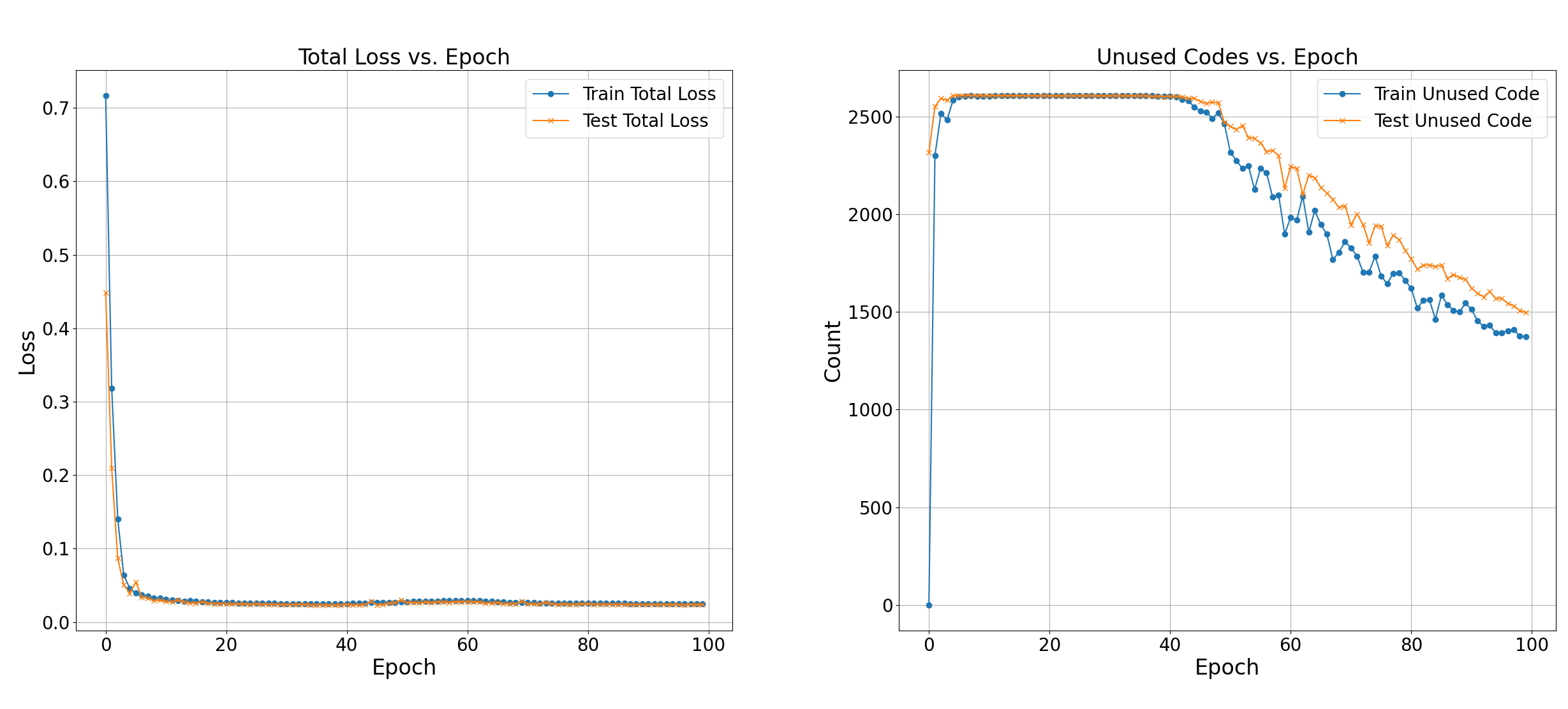}
    \caption{\textbf{Pre-training dynamics of the Tokenizer.} Left panel: convergence curves of the Total Loss for both training and testing sets. Right panel: trajectory of unused Codebook indices over epochs.}
    \label{fig:codebook_loss}
\end{figure}

In order to elucidate the training dynamics of the Stage 1 Spectrum-Aware Semantic Tokenizer, the total loss and codebook utilization were monitored throughout the pre-training process.Figure~\ref{fig:codebook_loss} shows the convergence curves for both training and validation sets, alongside the trajectory of unused codebook indices.

\textbf{Convergence and stability.} The left panel displays the evolution of the Tokenizer's total loss, $\mathcal{L}_{\text{Tokenizer}} = \mathcal{L}_{\text{Spec}} + \mathcal{L}_{\text{VQ}} + \mathcal{L}_{\text{Con}}$, which comprises the Spectral Reconstruction Loss, the Vector Quantization commitment loss, and the Semantic Consistency Loss.  During the initial training epochs, the loss decreases rapidly, indicating that the spectrum-aware reconstruction objective provides a strong inductive bias that enables the model to quickly capture the dominant quasi-periodic structures and fundamental frequency components of PPG signals. As training progresses, the loss plateaus and stabilizes. Notably, the close alignment between training and testing losses, with a consistently small generalization gap, suggests that the tokenizer learns a robust physiological signal manifold rather than overfitting to noise.

\textbf{Codebook evolution and utilization.} 
Figure~\ref{fig:codebook_loss}, right panel shows the evolution of unused codebook entries out of $K=4096$ 
during training. While many codes are initially inactive due to random initialization, their number steadily decreases and converges to a low level, indicating high codebook utilization.
This behavior supports two key conclusions. First, the steady reduction in unused codes indicates that the model avoids codebook collapse, effectively leveraging the full discrete space to represent fine-grained PPG morphological variations. Second, the high utilization rate validates the choice of 
$K=4096$, suggesting that PPG morphology exhibits sufficient diversity to require a high-capacity codebook and that the semantic consistency constraint preserves this diversity while clustering physiologically similar waveforms.

\subsection{SIGMA-PPG pre-training}
\begin{figure}[!h]
    \centering
    \includegraphics[width=\linewidth]{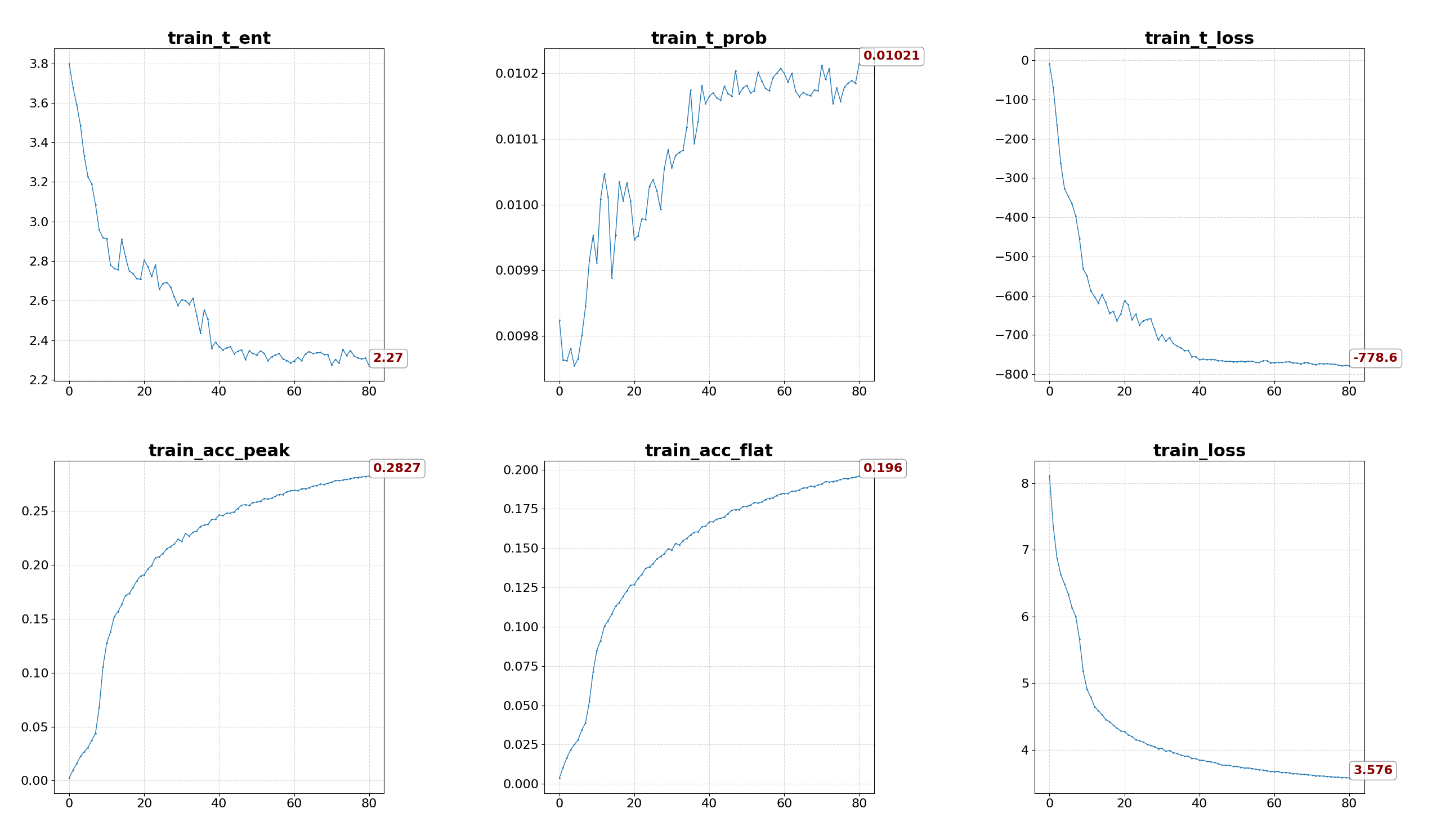}
    \caption{\textbf{Pre-training dynamics of the Tokenizer.} Top Row (Teacher Metrics): (1) train\_t\_loss decreases significantly, indicating the Teacher effectively maximizes the reward (Student error); (2) train\_t\_ent declines smoothly, marking a transition from random exploration to prior-guided exploitation; (3) train\_t\_prob increases slightly, reflecting growing confidence in mask selection. Bottom Row (Student Metrics): (4) train\_acc\_flat saturates at a lower 0.19, confirming that the model prioritizes high-information peaks over noisy flat regions; (5) train\_acc\_peak stabilizes around 0.28, demonstrating robust learning of morphological semantics; (6) train\_loss shows convergence despite increasing task difficulty.}
    \label{fig:pretrain_loss}
\end{figure}

We present a detailed analysis of the interaction dynamics between the Teacher and Student networks during SIGMA-PPG pre-training. Figure~\ref{fig:pretrain_loss} depicts the learning trajectories of six key metrics, highlighting how the Prior-Guided Adversarial Masking mechanism induces a physiological curriculum and effectively prevents representation collapse.

\subsubsection{Teacher Network Dynamics}


The Teacher network is trained to generate adversarial yet physiologically plausible masking patterns that maximize the Student’s reconstruction error under statistical prior constraints. During pre-training, the Teacher loss (train\_t\_loss) decreases sharply, indicating a substantial increase in reward and confirming that the Teacher progressively discovers more challenging and informative masking strategies. At the same time, the entropy of the Teacher’s policy (train\_t\_ent), $\mathcal{H}(\pi_\theta) = - \sum \pi_\theta(M_i) \log \pi_\theta(M_i)$, declines from a high initial value and stabilizes at a non-zero level, reflecting a controlled transition from broad exploration to focused exploitation without collapsing to a single masking pattern. Consistently, the average sampling probability (train\_t\_prob), defined as $\frac{1}{N} \sum \pi_\theta(M_i | x_i)$, increases slightly, showing growing confidence in masking physiologically salient regions. Together, these trends demonstrate that the Teacher converges to a stable, targeted masking policy that continuously challenges the Student while maintaining sufficient diversity to support robust representation learning.

\subsubsection{Student Network Learning Trajectory}

The function of the Student Network (Transformer Encoder) is to reconstruct the original semantic tokens from the masked input sequence. Its metrics reflect its adaptation to the dynamic difficulty imposed by the Teacher.

\paragraph{Student Reconstruction Loss.}
The Student optimizes the Cross-Entropy Loss (train\_loss)for masked tokens: $\mathcal{L}_{Student} = -\sum_{i=1}^{N} M_i \log p(z_i | \tilde{Z})$.
The curve shows a clear fast-then-slow convergence pattern. The initial sharp drop reflects the Student’s rapid learning of basic PPG periodicity and low-frequency components, while the later slower, oscillatory phase indicates sustained adversarial pressure as the Teacher targets finer morphological features (e.g., dicrotic notches), preventing trivial reconstruction strategies.
%

\paragraph{Peak vs. Flat reconstruction accuracy.}
The Student’s reconstruction performance differs by PPG signal region. We evaluate high-amplitude, high-skewness regions and low-amplitude, low-skewness regions. Accuracy on informative systolic peaks (train\_acc\_peak)) steadily increases and stabilizes around 0.28, indicating robust learning of meaningful pulse-wave morphology despite adversarial masking. In contrast, accuracy on low-information flat regions (train\_acc\_flat) remains lower, suggesting these segments are dominated by noise and limited structure. This confirms that SIGMA-PPG prioritizes physiologically relevant features over redundant baseline signals.


\section{Effectiveness of spectrum reconstruction}
\label{sec:spectrum}
To validate the Spectrum-Aware Tokenizer, we performed an ablation study on reconstruction objectives. Motivated by prior works (e.g., LaBraM), we compared four differt objectives: amplitude spectrum (Amp.), raw waveform (Raw), phase spectrum (Phase), and both amplitude and phase spectra.
As shown in Table \ref{tab:recon}, a counter-intuitive result emerges: incorporating phase information consistently degrades performance. The amplitude spectrum objective achieves the best results across most tasks, particularly for noise-sensitive regression (e.g., SpO2 MAE=4.557, SBP MAE=13.07). In contrast, phase spectrum objective alone performs poorly, and combining phase with amplitude generally reduces accuracy (e.g., Stress AUC drops from 0.9083 to 0.8948).



\begin{table}[h]
\centering
\caption{\textbf{Ablation study on different reconstruction objectives.} We compare the impact of reconstructing amplitude spectrum, (Amp.), raw signal (Raw), phase spectrum (Phase), and both spectra (Amp.+ Phase) on downstream task performance. Best results are highlighted in bold.}
\label{tab:recon}
\begin{small}
\begin{tabular}{lcccccc}
\toprule
\textbf{Classification} & Stress & Affect & Hyper & Signal Quality & Activity & Human Ident. \\
\midrule
Amp. & \textbf{0.9083} & 0.7702 & \textbf{0.7655} & \textbf{0.9601} & \textbf{0.8114} & \textbf{0.8906} \\
Raw & 0.8790 & \textbf{0.7720} & 0.7573 & 0.9421 & 0.7746 & 0.8502 \\
Phase & 0.8620 & 0.7124 & 0.7249 & 0.8948 & 0.7692 & 0.8416 \\
Amp.+ Phase & 0.8948 & 0.7675 & 0.7569 & 0.9505 & 0.8072 & 0.8762 \\
\midrule
\textbf{Regression} & RR & HR & SpO2 & SBP & DBP & Average HR \\
\midrule
Amp. & \textbf{1.880} & 5.831 & \textbf{4.557} & \textbf{13.07} & \textbf{7.748} & \textbf{6.729}\\
Raw & 1.933 & 6.779 & 4.566 & 13.266 & 7.903 & 6.756 \\
Phase & 2.102 & 6.193 & 4.990 & 14.168 & 8.174 & 7.536 \\
Amp.+ Phase & 1.917 & \textbf{5.784} & 4.648 & 13.318 & 7.891 & 6.831 \\
\bottomrule
\end{tabular}
\end{small}
\end{table}

\begin{figure}[!h]
    \centering
    \includegraphics[width=\linewidth]{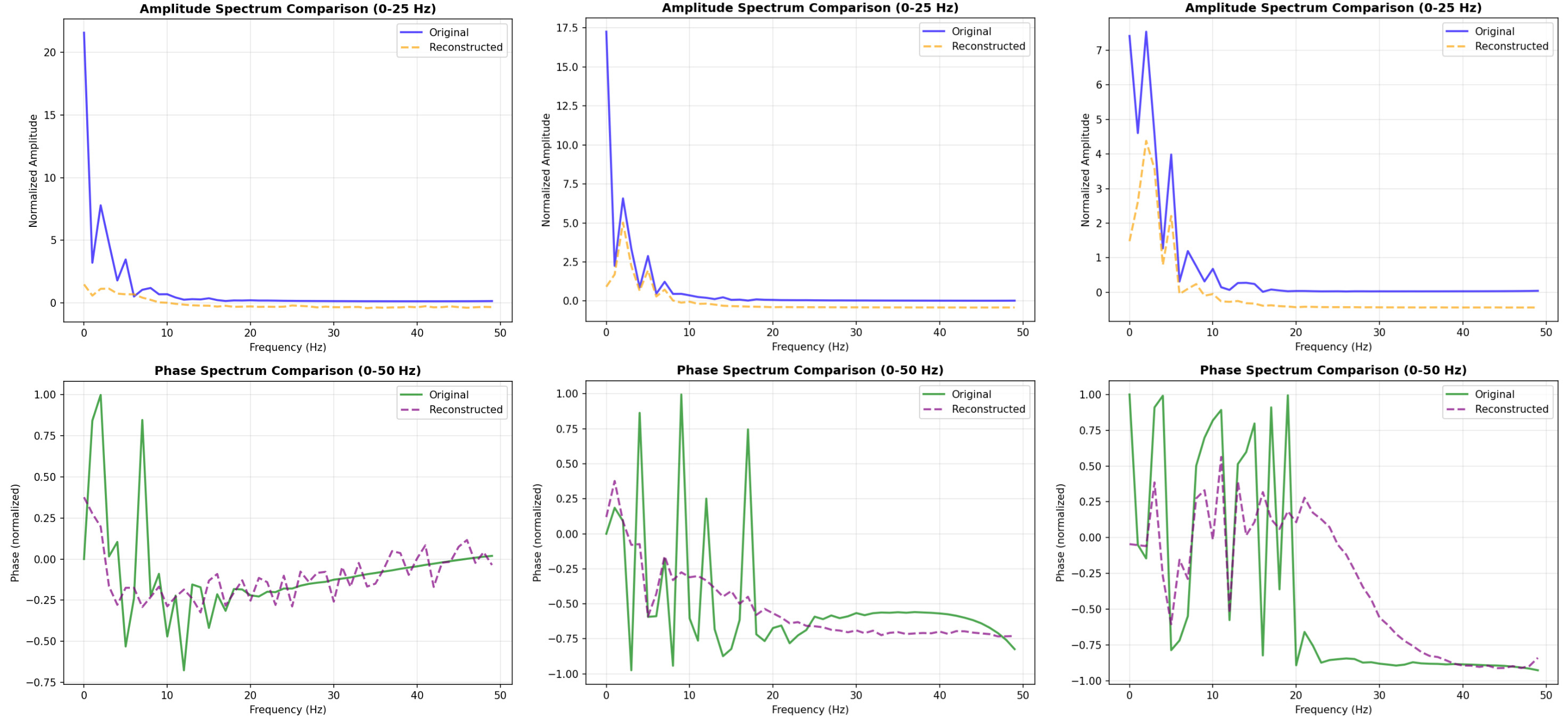}
    \caption{\textbf{Reconstruction quality across different training stages (epochs 0, 40, and 80).} The top row shows that the model with the amplitude spectrum objective converges rapidly to a high-fidelity reconstruction. In contrast, the bottom row shows that the model with the phase spectrum objective remains noisy and fails to converge.}
    \label{fig:recon}
\end{figure}

To investigate the source of this performance degradation, we visualized reconstructions at different training stages (epochs 0, 40, and 80), as shown in Figure \ref{fig:recon}. While the amplitude spectrum (top row) is recovered with high fidelity, the phase spectrum (bottom row) remains noisy and poorly reconstructed even at convergence. 
Phase reconstruction fails for two reasons: it is highly sensitive to temporal misalignment and to erratic noise. In our patch-based tokenization, the starting position of a patch is not related to the cardiac cycle phases, making it physiologically uninformative. PPG signal may be affected by random noise with a stochastic phase spectrum. As a result, reconstructing phase forces the model to learn irrelevant or noisy variations, degrading semantic representation quality.



Finally, we observed that reconstructing the raw PPG patch also yields suboptimal results compared to those obtained by using the amplitude spectrum. Raw waveform reconstruction relies on Mean Squared Error (MSE) in the time domain, which is notoriously sensitive to outliers and random noise that are difficult to recover (e.g., sensor shifts or baseline wander). In contrast, the amplitude spectrum is inherently shift-invariant and summarizes the signal's energy distribution. By focusing on the dominant frequency components, the amplitude objective acts as a natural denoising filter, encouraging the tokenizer to capture the "physiological essence" rather than the noisy morphology \citep{chen5473270instantaneous}.

\section{Ablation study on pre-training data source}
\label{sec:pretraining data}

\begin{table*}[!h]
\centering
\caption{\textbf{Performance comparison under the Linear Probing setting to isolate the impact of pre-training data.} Pulse-PPG denotes the official baseline weights. Retrained Pulse-PPG denotes the Pulse-PPG model trained from scratch using exactly the same pre-training dataset as our method to ensure a fair comparison of model architectures. Best results between the retrained baseline and our model are highlighted in \textbf{bold}.}
\label{tab:ablation_data_vs_arch}
\resizebox{\textwidth}{!}{%
\begin{tabular}{l|cccccc|cccccc}
\toprule
\multirow{2}{*}{\textbf{Method}} & \multicolumn{6}{c|}{\textbf{Regression (MAE $\downarrow$)}} & \multicolumn{6}{c}{\textbf{Classification (AUC $\uparrow$)}} \\
\cmidrule(lr){2-7} \cmidrule(lr){8-13}
 & RR & HR & SpO2 & SBP & DBP & AvgHR & Stress & Affect & Hyper. & Sig. Qual. & Activity & Ident. \\
\midrule
Pulse-PPG (Original) & 2.054 & 6.946 & 5.185 & 13.62 & 8.878 & \textbf{4.003} & \textbf{0.9687} & \textbf{0.7807} & \textbf{0.7971} & 0.9514 & 0.8051 & \textbf{0.9892} \\
\midrule
Retrained Pulse-PPG & 2.536 & 6.133 & 5.587 & 13.66 & 8.606 & 4.759 & 0.8670 & 0.7591 & \textbf{0.7740} & 0.9562 & 0.7698 & 0.9815 \\
\textbf{SIGMA-PPG (Ours)} & \textbf{1.880} & \textbf{5.831} & \textbf{4.557} & \textbf{13.07} & \textbf{7.748} & 6.729 & \textbf{0.9083} & \textbf{0.7702} & 0.7655 & \textbf{0.9601} & \textbf{0.8114} & 0.8906 \\
\bottomrule
\end{tabular}%
}
\end{table*}

As shown in Table~\ref{tab:ablation_data_vs_arch}, retraining Pulse-PPG on the same clinical dataset used for SIGMA-PPG results in a marked performance drop across most tasks. In contrast, SIGMA-PPG consistently achieves lower errors on morphology-sensitive regression problems, including an 18.4\% reduction in SpO2 MAE and a 4.3\% reduction in SBP MAE relative to the retrained Pulse-PPG. These results indicate that the performance gains arise from the generative masked modeling architecture rather than differences in data scale. While contrastive objectives used in Pulse-PPG excel at global identity features (mean HR, and human identification), our statistical-prior guided generative approach better captures finer-grained hemodynamic structure, leading to more accurate physiological estimation and improved robustness under domain shift.

\section{Sensitivity analysis of statistical prior weights}
\label{sec:beta}

\begin{table*}[!h]
    \centering
    \caption{\textbf{Sensitivity analysis of the hyperparameter $\beta$.} It balances the contribution of Amplitude Stability ($S_{amp}$) and Morphological Skewness ($S_{skew}$) in the statistical prior scoring function. The best results are highlighted in \textbf{bold}.}
    \label{tab:beta_sensitivity}
    
    \begin{subtable}{\linewidth}
        \centering
        \caption{Regression Tasks (MAE $\downarrow$)}
        \setlength{\tabcolsep}{6pt}
        \begin{tabular}{lcccccc}
            \toprule
            $\beta$ & RR & HR & SpO2 & SBP & DBP & AvgHR \\
            \midrule
            1.0 (Pure Skewness) & 1.949 & 6.028 & 4.842 & 14.54 & 8.757 & 7.223 \\
            0.7 & 1.920 & 5.983 & \textbf{4.014} & 13.53 & 7.785 & 7.101 \\
            \textbf{0.5 (Balanced)} & \textbf{1.880} & 5.831 & 4.557 & \textbf{13.07} & \textbf{7.748} & \textbf{6.729} \\
            0.3 & 1.932 & \textbf{5.414} & 4.862 & 13.76 & 7.844 & 7.355 \\
            0.0 (Pure Amplitude) & 1.950 & 6.562 & 4.962 & 14.60 & 8.694 & 7.566 \\
            \bottomrule
        \end{tabular}
    \end{subtable}
    
    \vspace{1em} 

    \begin{subtable}{\linewidth}
        \centering
        \caption{Classification Tasks (AUC $\uparrow$)}
        \setlength{\tabcolsep}{5pt}
        \begin{tabular}{lcccccc}
            \toprule
            $\beta$ & Stress & Affect & Hyper. & Sig. Quality & Activity & Human Ident. \\
            \midrule
            1.0 (Pure Skewness) & \textbf{0.9166} & 0.7698 & 0.7567 & 0.9415 & 0.8008 & 0.8897 \\
            0.7 & 0.9099 & \textbf{0.7868} & 0.7643 & 0.9520 & 0.8038 & \textbf{0.8939} \\
            \textbf{0.5 (Balanced)} & 0.9083 & 0.7702 & \textbf{0.7655} & \textbf{0.9601} & 0.8114 & 0.8906 \\
            0.3 & 0.8999 & 0.7776 & 0.7426 & 0.9540 & \textbf{0.8133} & 0.8912 \\
            0.0 (Pure Amplitude) & 0.8854 & 0.7508 & 0.7413 & 0.9495 & 0.7825 & 0.8839 \\
            \bottomrule
        \end{tabular}
    \end{subtable}
\end{table*}

We performed a sensitivity analysis on the weighting parameter $\beta$ to assess the relative contributions of Amplitude Stability ($S_{\text{amp}}$) and Morphological Skewness ($S_{\text{skew}}$) (Table~\ref{tab:beta_sensitivity}).

\textbf{Regression}. A balanced combination performs best: $\beta=0.5$ yields the lowest MAE on key cardiovascular tasks (e.g., SBP 13.07, DBP 7.75, RR 1.88), whereas extreme settings ($\beta=0.0$ or $1.0$) consistently increase errors, indicating that neither prior alone is sufficient.

\textbf{Classification}. The same trend holds overall. While pure skewness ($\beta=1.0$) slightly improves Stress detection, $\beta=0.5$ achieves the highest AUC for Signal Quality (0.9601) and Hypertension (0.7655).

In onclusion, Amplitude Stability and Skewness provide complementary cues—capturing signal integrity and pulse morphology, respectively. We therefore adopt $\beta=0.5$ as the default setting for SIGMA-PPG to ensure robust, task-agnostic performance.

\section{Related work}

\subsection{PPG foundation model}

Pioneering works such as PaPaGei \citep{pillai2024papagei}, Pulse-ppg\citep{saha2025pulse}, AnyPPG \citep{nie2025anyppg}, and QualityFM \citep{guo2025qualityfm} have successfully adapted contrastive frameworks to the physiological domain. By maximizing the similarity between augmented views or cross-modal pairs (e.g., PPG and ECG), these models learn robust, noise-invariant representations. However, the discriminative nature of contrastive objectives inherently encourages the model to learn global semantic invariance (e.g., subject identity or activity state) at the expense of fine-grained morphological precision. Crucial hemodynamic details—such as the dicrotic notch or slope variations required for dense regression tasks (e.g., blood pressure)—are often suppressed as "intra-sample noise" or nuisance variability, limiting their performance on tasks requiring precise waveform reconstruction.

To address the lack of local granularity, generative masked modeling has been introduced, with methods like HiMAE \citep{lee2025himae} and MMR \citep{thukral2025wavelet} reconstructing raw signals from partial observations. Nevertheless, applying standard random masking to PPG signals faces a unique challenge: intrinsic redundancy. Due to the high quasi-periodicity of cardiac cycles, models can easily minimize training loss by performing trivial local interpolation from adjacent beats without learning meaningful semantic structures. This leads to representations that capture low-level periodicity but fail to encode the underlying physiological drivers.

Alternatively, GPT-PPG \citep{radford2018improving} adopts an autoregressive (AR) approach, tokenizing signals and predicting future values in a sequential manner. While this paradigm aligns well with forecasting and regression tasks due to its focus on temporal continuity, it often yields suboptimal results in classification scenarios.

\subsection{Adversarial masking and curriculum learning}

The paradigm of masked modeling, pioneered by BERT \citep{devlin2019bert} in NLP and MAE \citep{he2022masked} in Computer Vision, has established itself as a powerful proxy task for representation learning. By reconstructing masked tokens from partial observations, these models learn robust contextual dependencies. However, standard uniform random masking employed in these frameworks often suffers from inefficiency.

To address this, Curriculum Learning \citep{bengio2009curriculum} and Adversarial Masking \citep{shi2022adversarial} have been introduced to optimize the training process. Instead of static random sampling, these approaches formulate mask generation as a minimax game. A "Teacher" network learns to dynamically generate challenging masking patterns that maximize the "Student's" reconstruction loss \citep{williams1992simple}. This mechanism effectively creates an adaptive curriculum, compelling the learner to move beyond simple local interpolation and capture high-level structural semantics.

\subsection{Discrete Representation and Vector Quantization}

Learning discrete latent representations has proven effective in capturing high-level semantic abstractions while filtering out low-level noise. The seminal work of VQ-VAE \citep{van2017neural} pioneered the mapping of continuous data into a fixed discrete codebook. To further enhance modeling capacity and capture long-range dependencies, subsequent studies have developed hierarchical and multi-scale frameworks. For instance, VQ-VAE-2 \citep{razavi2019generating} employs multi-level latent maps to generate high-fidelity samples, while Jukebox \citep{dhariwal2020jukebox} leverages hierarchical VQ to model raw audio waveforms across varying timescales. Additionally, VQ-GAN \citep{esser2021taming} integrates adversarial feedback to significantly improve the expressivity and perceptual quality of the codebook. However, the inherent instability of the discretization process when applied to continuous physiological time-series cannot be ignored.

\section{Final remarks}
Our results position SIGMA-PPG as a robust framework for physiological representation learning, effectively separating core hemodynamics from signal redundancy.

\textbf{Temporal context.} Ablations show performance peaks with a 240-s input window, especially for regression (e.g., in the SpO2 estimation task). This duration spans hundreds of cardiac cycles enabling attention to average out transient artifacts and variability. Longer contexts are currently limited by quadratic attention costs, which efficient attention could alleviate.

\textbf{Limitations—domain shift.} A clear distribution gap exists between clinical pre-training data and those collected from wearable devices. Lower performance using Linear Probing on wrist-worn PPG datasets reflects morphological and noise differences, indicating that fine-tuning is still required for effective adaptation.

\textbf{Resource efficiency.} While powerful, SIGMA-PPG’s size challenges edge deployment. Future work will focus on compression—knowledge distillation, pruning, and quantization (e.g., from FP32 to INT8)—to enable low-power, on-device inference with improved privacy.

\subsection{Future directions}
To bridge the gap between clinical capability and ubiquitous utility, future research should prioritize:
\begin{itemize}
    \item \textbf{Daily life data.} Incorporating large-scale data from consumer wearables to enforce invariance to intense motion artifacts and environmental noise.
    \item \textbf{Domain Generalization.} Exploring techniques like adversarial training to align feature spaces between transmissive (clinical) and reflective (wearable) PPG signals.
    \item \textbf{Edge Adaptation.} Implementing the discussed model compression techniques (distillation, pruning, and quantization) to enable real-time, privacy-preserving deployment on low-power wearable devices.
\end{itemize}

In conclusion, SIGMA-PPG demonstrates the potential of generative foundation models in healthcare. Addressing current data limitations and scaling context and computational efficiency will be key to establishing a truly universal physiological encoder.

\end{document}